\documentclass[printer]{tRES2e}

\usepackage{color}
\usepackage{algorithm}
\usepackage{algorithmic}
\usepackage[below]{placeins}%
\usepackage{amsmath}%

\usepackage{amsfonts}%
\usepackage{amssymb}%
\usepackage{graphicx}
\expandafter\let\csname ver@natbib.sty\endcsname\relax
\begin{document}

\newcommand{\al}{\alpha}
\newcommand{\be}{\beta}
\newcommand{\la}{\lambda}
\newcommand{\f}{\mathbf{f}_{\lambda}}
\newcommand{\fr}{\widehat{\mathbf{f}}_{\lambda}}
\newcommand{\ca}{\mathbf{c}^{\mathbf{f}}_{\alpha}}
\newcommand{\cb}{\mathbf{c}^{\mathbf{p}}_{\beta}}

\markboth{G. Noyel \& J. Angulo and D. Jeulin}{Postprint: A new
spatio-spectral morphological segmentation for remote sensing}

\articletype{Postprint}

\title{A new spatio-spectral morphological segmentation for multispectral remote sensing images}
\author{G. NOYEL$^{\ast}$$\dag$\thanks{$^{\ast}$Corresponding author. Email:
noyel@cmm.ensmp.fr \vspace{6pt}} \& J. ANGULO${\dag}$ and D.
JEULIN${\dag}$\\\vspace{6pt}$\dag$MINES ParisTech, CMM - Centre de
Morphologie Math\'ematique, Math\'ematiques et Syst\`emes, 35 rue
Saint Honor\'e - 77305 Fontainebleau cedex, France\\} \maketitle

\begin{abstract}
A general framework of spatio-spectral segmentation for
multispectral images is introduced in this paper. The method is
based on classification-driven stochastic watershed by Monte Carlo
simulations, and it gives more regular and reliable contours than
standard watershed. The present approach is decomposed into several
sequential steps. First, a dimensionality reduction stage is
performed using Factor Correspondence Analysis method. In this
context, a new way to select the factor axes (eigenvectors)
according to their spatial information is introduced. Then a
spectral classification produces a spectral pre-segmentation of the
image. Subsequently, a probability density function (pdf) of
contours containing spatial and spectral information is estimated by
simulation using a stochastic watershed approach driven by the
spectral classification. The pdf of contours is finally segmented by
a watershed controlled by markers coming from a regularization of
the initial classification.

%Keywords: Stochastic watershed, multispectral images segmentation, probability
%density function of contours, spatio-spectral segmentation, mathematical morphology.

\end{abstract}

\section{Introduction}

Multispectral (or hyperspectral) images which are composed of
several tens or hundreds of spectral bands are nowadays used
currently in remote sensing. These spectral bands bring a lot of
information concerning the structure of the ground, the vegetation,
buildings, etc. However, in order to efficiently process these large
amount of data, one has to develop new methods to analyse these
images and especially to segment them (i.e. to group similar pixels
into connected classes). Standard methods of segmentation require to
be extended to this type of rich images.

Two types of information are carried out by multi/hyper-spectral
images:
\begin{enumerate}
\item spatial information contained in the position of each pixel and in the geometry of neighborhood pixels according to
the location on the bitmap grid;
\item spectral information contained in the spectral bands, in such a way that a spectrum is
associated to each pixel.
\end{enumerate}

Using simultaneously both kinds of information is a crucial point to
get a robust and accurate method of segmentation. More precisely,
one can compare the spectral information of each pixel and group
them into classes. This comparison is global on the image, since
each pixel is compared to all the others. However, in this
point-wise comparison, the spatial relationship is usually not taken
into account. Another method consists in comparing the pixels in a
neighborhood leading to an approach based on a spatial comparison,
but usually this spatial processing is achieved impenitently for
each spectral band. With these remarks, we understand the necessity
of combining both kinds of information.

\textbf{State-of-the-art on segmentation of multispectral images.}
In the literature, there are several other methods of segmentation
that can be used for segmentation of multispectral images.

Watershed based segmentation was used by \citet{Soille_1996} to
separate into classes the histogram of the multispectral image. Mean
Shift algorithm was used to segment hyperspectral data cubes by
\citet{Genova_2006}. Moreover, various classification methods were
considered to group pixels into non-connected classes. Support
Vector Machines (SVM) were applied for instance to remote sensing by
\citet{Gualtieri_1999,Lennon_2002_SVM,Archibald_2007}. One of the
advantages of this method is not to be sensitive to dimensionality
problem, also called ``Hughes phenomenon"
\citep{Hughes_1968,Landgrebe_2002}. ``$K$-nearest neighbours" was
used as a spatio-spectral classification method by
\citet{Marcal_2005,Polder_2004}. In \citet{Schmidt_2007}, a
classification method based on wavelets was introduced.

In order to analyse the structures in an image,
\citet{Pesaresi_2001} developed the morphological profile,
%(MP),Benediktsson_2003}
 based on granulometry principle
\citep{Serra_1982}. The Morphological Profile is composed of a
series of openings by reconstruction of increasing sizes and of a
series of closings by reconstruction (obtained by the dual
operation) \citep{Soille_1999}. With these operations, the size and
the shape of the objects contained in the image are determined.
Therefore, the MP contains only spatial information. A
straightforward method to extend MP to hyperspectral image, is to
build the MP for each channel. This approach is called Extended
Morphological Profile (EMP). In fact, \citet{Benediktsson_2005} use
the first principal components obtained by a Principal Component
Analysis of the hyperspectral image. Consequently, EMP contains
spatial and spectral information. Finally a classifier, such as a
SVM or a neural network \citep{Fauvel_2007}, is applied on EMP to
obtain a classification in which the classes are not necessarily
connected. In the present paper, we consider a segmentation as a
partition of connected classes of an image following the definition
given in \citet{Serra_2006} in the framework of lattice theory.

Gradient based approaches such as the watershed segmentation cannot
detect thin objects in remote sensing images. Indeed, thin objects
have no interior on the image of the gradient. Consequently, the
flooding procedure which defines a region starting from markers on
the gradient, does not take into account the thin objects. These
thin objects and other ``small'' structures can be obtained, on the
one hand, using segmentation techniques based on connective criteria
for pixel aggregation~\citep{Noyel_ISMM_2007, Soille_2008}, or on
the other hand, using the residue of an opening (known as top-hat
operator) which has been extended to multivariate images
in~\citet{AnguloSerra07}. As discussed in~\citet{Soille_2008}, this
is a limitation for some applications, e.g., extraction of road
networks in aerial images.

The watershed transformation (WS) is one of the most powerful tools
for segmenting images and was introduced by \citet{Beucher_1979}.
According to the flooding paradigm, the watershed lines associate a
catchment basin to each minimum of the relief to flood (i.e. a
greyscale image) \citep{Beucher_1992}. Typically, the relief to
flood is a gradient function which defines the transitions between
the regions. Using the watershed on a scalar image without any
preparation leads to a strong over-segmentation (due to a large
number of minima). There are two alternatives in order to get rid of
the over-segmentation. The first one consists in initially
determining markers for each region of interest. Then, using the
homotopy modification, the only local minima of the gradient
function are imposed by the region markers. The extraction of the
markers, especially for generic images, is a difficult task. The
second alternative involves hierarchical approaches based on
non-parametric merging of catchment basins (waterfall algorithm) or
based on the selection of the most significant minima according to
different criteria (dynamics, area or volume extinction values)
using extinction functions \citep{Meyer_2001}.

We have introduced a general method to segment hyperspectral images
by WS~\citep{Noyel_IAS_2007}. Several multivariate gradients were
studied. The markers are obtained from a previous spectral
classification.

Although this approach is powerful, it is a deterministic process
which tends to build irregulars contours and can produce an over-segmentation.
The stochastic WS was proposed in order to regularize
and to produce more significant contours
\citep{AnguloJeulin_ISMM_2007}. The initial framework was then
extended to multispectral images
\citep{Noyel_KES_2007,Noyel_CGIV_2008,Noyel_PhD_2008}.

\textbf{Aim of the paper.} Based on the experience of our previous
works on multivariate image segmentation, the aim of this study is
to present a complete chain for an automatic spatio-spectral
segmentation of multispectral images, and to illustrate the method
with some examples of remote sensing images.

\begin{figure}
\begin{center}
\includegraphics[width=0.9\columnwidth]{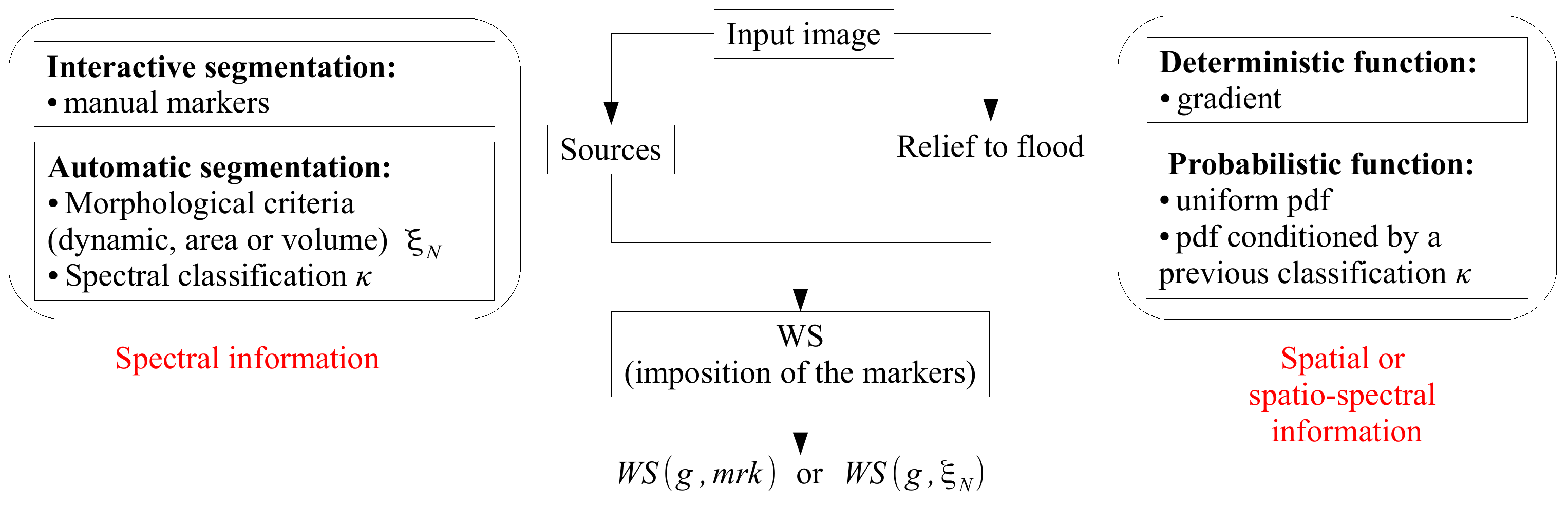}\newline
\end{center}
\caption{General framework of multispectral image segmentation}%
\label{fig_WS_framework}%
\end{figure}

Let us start by presenting a general paradigm of WS-based
segmentation of multispectral images (fig. \ref{fig_WS_framework}).
This segmentation requires two basic ingredients:
\begin{enumerate}
\item some markers for the regions of interest $mrk$;
\item a relief to flood which describes the ``energy'' of the frontiers between the regions $g$.
\end{enumerate}
The markers can be chosen interactively by a user or automatically
by means of a morphological criteria $\xi_{N}$, or with the classes
of a previous spectral classification. The relief to flood is a
scalar function (i.e., a greyscale image). For the deterministic WS,
it is usually a gradient (in fact its norm), or a distance function.
For the stochastic WS, the function to flood is a probability
density function (pdf) of the contours appearing in the image. After
extracting the markers, they are imposed as sources of the relief to
flood and the WS is computed. The results are noted $WS(g,mrk)$ or
$WS(g,\xi_{N})$.

\bigskip

The segmentation framework for remote sensing images introduced in
this paper is coherent with the general paradigm of WS-based
segmentation. More precisely, the method is based on
classification-driven stochastic watershed by Monte Carlo
simulations, and as we will show, it gives more regular and reliable
contours than standard watershed.

The approach is decomposed into several sequential steps. Each step
was partially considered in our previous works, where our methods
were compared with more standard ones. However, the main objective
of this paper is to introduce the global pipeline, presenting
exclusively for each step the algorithm leading to the best
performance.

Firstly in section \ref{sec_dim_red}, a dimensionality reduction
stage is performed using Factor Correspondence Analysis (FCA)
\citep{Benzecri_1973}. In this context, a new way to select the
factor axes (i.e. the eigenvectors) according to their spatial
information is introduced. Then a spectral classification on the
factor space, described in section
\ref{sec_generation_markers_classification}, produces a spectral
pre-segmentation of the image. Subsequently, a probability density
function (pdf) of contours containing spatial and spectral
information is estimated by simulation using a stochastic WS
approach driven by the spectral classification. In section
\ref{sec_intro_sto_WS}, the standard stochastic WS is reminded, and
then, in section \ref{sec_seg_sto_WS}, the algorithm for the
construction of the classification-driven stochastic WS is
discussed. It also includes how the pdf of contours is finally
segmented by a watershed controlled by markers coming from a
regularization of the initial classification.

The paper is completed with section \ref{sec_notations}, which fixes
the notations used in this paper and which presents the image data
considered in the experiments. Finally, in section
\ref{sec_conclusion}, the conclusions and perspectives of the paper
are discussed.

\FloatBarrier
\section{Notations}
\label{sec_notations}

In a formal way, each pixel of a multispectral image is a vector
with values in wavelength. To each wavelength corresponds an image
in two dimensions called channel. The number of channels depends on
the nature of the specific problem under studies (satellite imaging,
spectroscopic images, temporal series, etc.).  Let
\begin{equation}
\mathbf{f_{\lambda}}:\left\{
\begin{array}
[c]{lll}%
E & \rightarrow & \mathcal{T}^{L}\\
x & \rightarrow & \mathbf{f}_{\mathbf{\lambda}}(x)=\left(  f_{\lambda_{1}%
}(x),f_{\lambda_{2}}(x),\ldots,f_{\lambda_{L}}(x)\right)  \\
\end{array}
\right.  \label{eq_im}%
\end{equation}
denote an hyperspectral image, where:

\begin{enumerate}

\item[$\bullet$] $E \subset\mathbb{R}^{2}$, $\mathcal{T} \subset\mathbb{R}$
and $\mathcal{T}^{L} = \mathcal{T} \times\mathcal{T} \times\ldots
\times\mathcal{T}$

\item[$\bullet$] $x = x_{i} \ /\ i\in\{1,2, \ldots, P \}$ is the
spatial coordinate of a vector pixel $\mathbf{f}_{\lambda}(x_{i})$
($P$ is the number of pixels of $E$)

\item[$\bullet$] $f_{\lambda_{j}} \ /\ j \in\{1,2, \ldots, L\}$ is a
channel ($L$ is the number of channels)

\item[$\bullet$] $f_{\lambda_{j}}(x_{i})$ is the value of vector pixel
$\mathbf{f}_{\lambda}(x_{i})$ on channel $f_{\lambda_{j}}$.
\end{enumerate}

\begin{figure}
\centering
\begin{tabular}
[c]{@{}c@{ }c@{ }c@{}}%
\includegraphics[width=0.33\columnwidth]{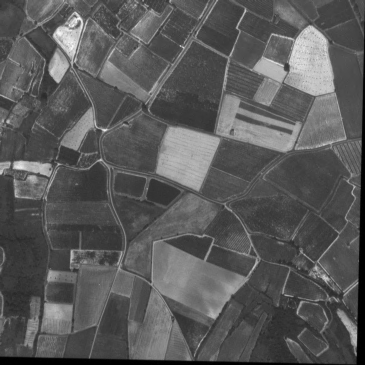} &
\includegraphics[width=0.33\columnwidth]{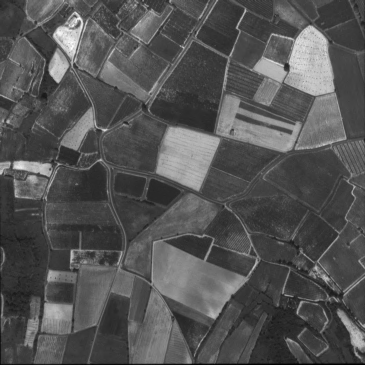} &
\includegraphics[width=0.33\columnwidth]{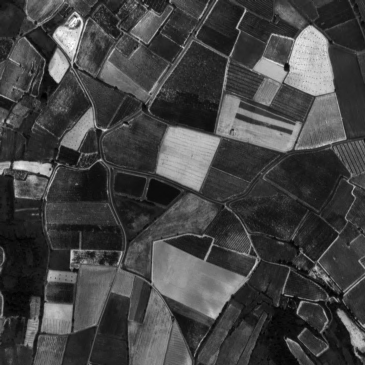}\\
{\small \emph{(a)} $\mathbf{f}_{\lambda_{1}}$} & {\small \emph{(b)}
$\mathbf{f}_{\lambda_{2}}$} &
{\small \emph{(c)} $\mathbf{f}_{\lambda_{3}}$}\\
\includegraphics[width=0.33\columnwidth]{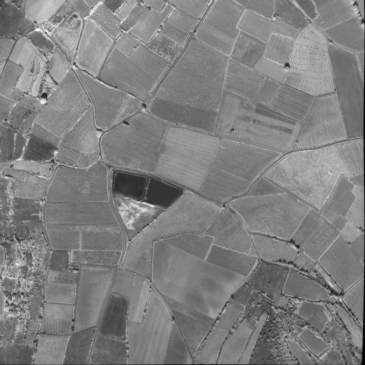} &
\includegraphics[width=0.33\columnwidth]{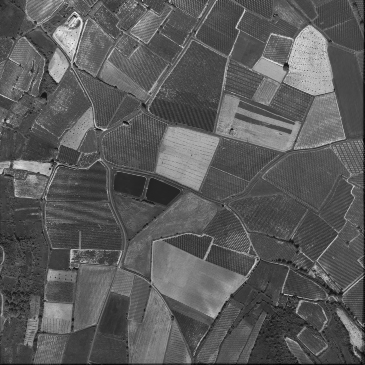} &
\includegraphics[width=0.33\columnwidth]{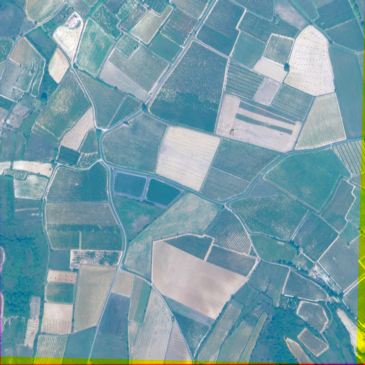}\\
{\small \emph{(d)} $\mathbf{f}_{\lambda_{4}}$} & {\small \emph{(e)}
$\mathbf{f}_{\lambda_{5}}$} &
{\small \emph{(f)} RGB}\\
&  &
\end{tabular}
\caption{Channels of multispectral image
$\mathbf{f}_{\mathbf{\lambda}}$ ``Roujan" (\copyright CNES):
\emph{(a)} $\mathbf{f}_{\lambda_{1}}$ blue, \emph{(b)}
$\mathbf{f}_{\lambda_{2}}$ green, \emph{(c)}
$\mathbf{f}_{\lambda_{3}}$ red, \emph{(d)}
$\mathbf{f}_{\lambda_{4}}$ near infra-red, \emph{(e)}
$\mathbf{f}_{\lambda_{5}}$ panchromatic,
\emph{(f)} synthetic RGB representation.}%
\label{fig_im_roujan}%
\end{figure}

Figure \ref{fig_im_roujan} gives an example of a five band satellite
simulated image PLEIADES, acquired by the CNES (Centre National
d'Etudes Spatiales, the French space agency) and provided by
\citet{Flouzat_1998}. Its channels are the following:
$f_{\lambda_{1}}$ blue, $f_{\lambda_{2}}$ green, $f_{\lambda_{3}}$
red, $f_{\lambda_{4}}$ near infrared and $f_{\lambda_{5}}$
panchromatic. The panchromatic channel, initially $1460\times1460$
pixels with a resolution of 0.70 meters, was resized to $365
\times365$ pixels. Therefore, the resolution is 2.80 meters in an
image of $365 \times365 \times5$ pixels. In order to represent a
multispectral image in a synthetic way, we created a synthetic RGB
image using channels $f_{\lambda_{3}}$ red, $f_{\lambda_{2}}$ green
and $f_{\lambda_{1}}$ blue.

The downsampling method consists in, by averaging, keeping only one
pixel on four pixels in both directions of the plane. The
coordinates $(i,j)$ of the pixels are ordered as the standard
ordering used for matrices. Moreover, we also notice that even if
the downsampled panchromatic image has values strongly correlated to
average intensity of the red, blue and green channels, the spatial
resolution of the panchromatic is better than the chromatic bands,
and consequently there is a better resolution of the contours which
is important for the aim of image segmentation. Indeed, to deal with
the redundancy, the FCA (Factor Correspondence Analysis) transforms
the original channels into uncorrelated factor channels (according
to the chi-squared metric).

\FloatBarrier
\section{Dimensionality reduction}
\label{sec_dim_red}

As the number of channels, in a multispectral image, is usually
important, the spectrum information contained in these channels is
redundant. Therefore, in order to avoid the Hughes phenomenon
\citep{Hughes_1968,Landgrebe_2002} associated to the spectral
dimensionality problem as well as to reduce the computational time,
it is necessary to reduce the number of channels.

Consequently, a data reduction is performed using Factor
Correspondence Analysis (FCA) \citet{Benzecri_1973}. We prefer using
a FCA instead of a Principal Component Analysis (PCA) or a Maximum
Noise Fraction (MNF) \citep{Green_1988}, because image values are
positive and the spectral channels can be considered as probability
distributions.

Green et al. 1988 have demonstrated that PCA and MNF are equivalent
in the case of an uncorrelated noise with equal variance in all
channels. This explains why PCA is often used to remove noise on
hyperspectral images in which the noise is almost uncorrelated with
the same variance on all channels.

As for PCA, from selected factorial FCA axes the image can be
partially reconstructed. The metric used in FCA is the chi-squared,
which is adapted to probability laws and normalized by channels
weights. FCA can be seen as a transformation going from image space
to a factorial space. In the factorial space, the coordinates of the
pixel vector, on each factorial axis, are called pixel factors. The
pixel factors can be considered as another multispectral image whose
channels correspond to the factorial axes:

\begin{equation}
\label{eq_FCA}\zeta: \left\{
\begin{array}
[c]{lll}%
\mathcal{T}^{L} & \rightarrow & \mathcal{T}^{K} \text{ / } K < L\\
\mathbf{f}_{\mathbf{\lambda}}(x) & \rightarrow & \mathbf{c}^{\mathbf{f}%
}_{\alpha}(x) = \left(  c^{\mathbf{f}}_{\alpha_{1}}(x), \ldots, c^{\mathbf{f}%
}_{\alpha_{K}}(x) \right) \\
\end{array}
\right.
\end{equation}

\begin{figure}
\centering
\begin{tabular}
[c]{@{}c@{ }c@{}}%
\includegraphics[width=0.33\columnwidth]{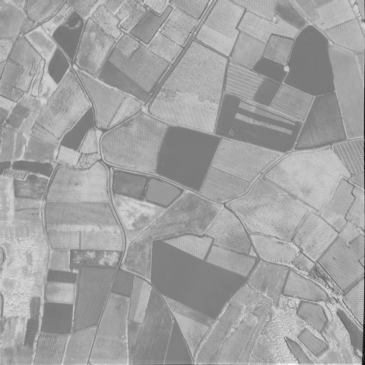} &
\includegraphics[width=0.33\columnwidth]{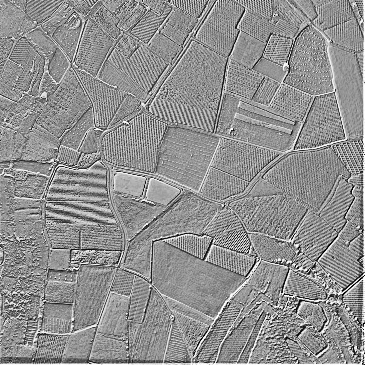}\\
{\small \emph{(a)} $c^{\mathbf{f}}_{\alpha_{1}}$} & {\small \emph{(b)} $c^{\mathbf{f}}_{\alpha_{2}}%
$}\\
\includegraphics[width=0.33\columnwidth]{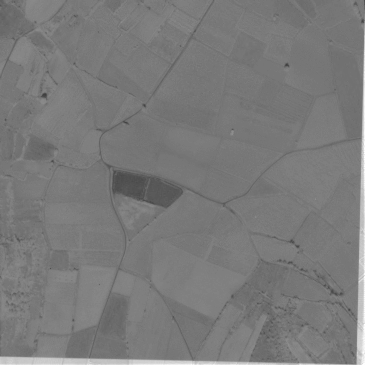} &
\includegraphics[width=0.33\columnwidth]{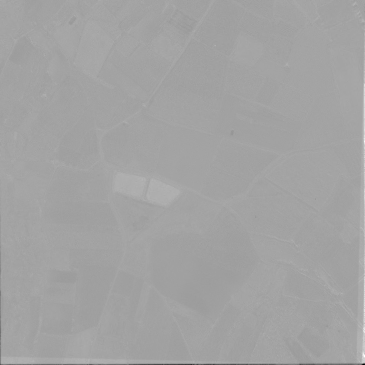}\\
{\small \emph{(c)} $c^{\mathbf{f}}_{\alpha_{3}}$} & {\small \emph{(d)} $c^{\mathbf{f}}_{\alpha_{4}}%
$}\\
&
\end{tabular}
\caption{FCA factors pixels $\mathbf{c}^{\mathbf{f}}_{\alpha}$ of
image ``Roujan" on axes \emph{(a)} 1, \emph{(b)} 2, \emph{(c)} 3 and
\emph{(d)} 4 with respective inertias 84.1 \%, 8.7 \%, 6.2
\%, 1 \%.}%
\label{fig_axes_roujan}%
\end{figure}

A limited number $K$, with $K<L$, of factorial axes is usually
chosen. Therefore FCA can be seen as a projection of the initial
vector pixels in a factor space with a lower dimension. Moreover, as
for PCA, FCA reduces the spectral noise on multispectral images
\citep{Green_1988,Noyel_IAS_2007,Noyel_PhD_2008}. Figure
\ref{fig_axes_roujan} shows FCA factor pixels
$\mathbf{c}_{\alpha}^{\mathbf{f}}$ of image ``Roujan". Consequently,
we have two spaces for multivariate segmentation: the multispectral
image space (MIS) and the factor image space (FIS).

A common problem in remote sensing application is the registering of
the various spectral images. If the registering step is not perfect,
the spatial shifts introduce in the dimensionality reduction a FCA
factor containing this information; see for instance, in figure 3,
the FCA factor image $c_{\alpha_2}^{\mathbf{f}}$. This image looks
like a laplacian image, but for the purpose of segmentation, it
should be considered as a factor image containing spatial noise. We
only show below how to detect and remove these ``noisy'' factor
images.

Another approach to reduce the number of channels consists in modelling the
spectrum of each vector-pixel $\mathbf{f}_{\lambda}(x_{i})$ and creating a new
multispectral image composed of the parameters of the model
\citep{Noyel_IAS_2007}; however, for generic multispectral remote sensing
images, this kind of modelling is quite difficult.

As shown by \citet{Benzecri_1973} and by \citet{Green_1988}, some
factor pixels on factor axes contain mainly noise, and others signal
information. In order to select relevant axes containing information
(i.e. signal against noise), we introduce a new method based on a
measurement of the signal to noise ratio $SNR$ of every factor
image. For each factorial axis $\alpha_{k}$, the centered spatial
covariance is computed by a 2D FFT (Fast Fourier Transform) on the
pixel factors:
\[
\overline{g}_{\alpha_{k}}(h)=E[\overline{c}_{\alpha_{k}}^{\mathbf{f}%
}(x)\overline{c}_{\alpha_{k}}^{\mathbf{f}}(x+h)],
\]
with $\overline{c}_{\alpha_{k}}^{\mathbf{f}}(x)=c_{\alpha_{k}}^{\mathbf{f}%
}(x)-E[c_{\alpha_{k}}^{\mathbf{f}}(x)]$, where $E$ is the mathematical
expectation (here the statistical expectation).

\begin{figure}
\centering
\begin{tabular}
[c]{@{}c@{ }c@{}}%
\includegraphics[width=0.5\columnwidth]{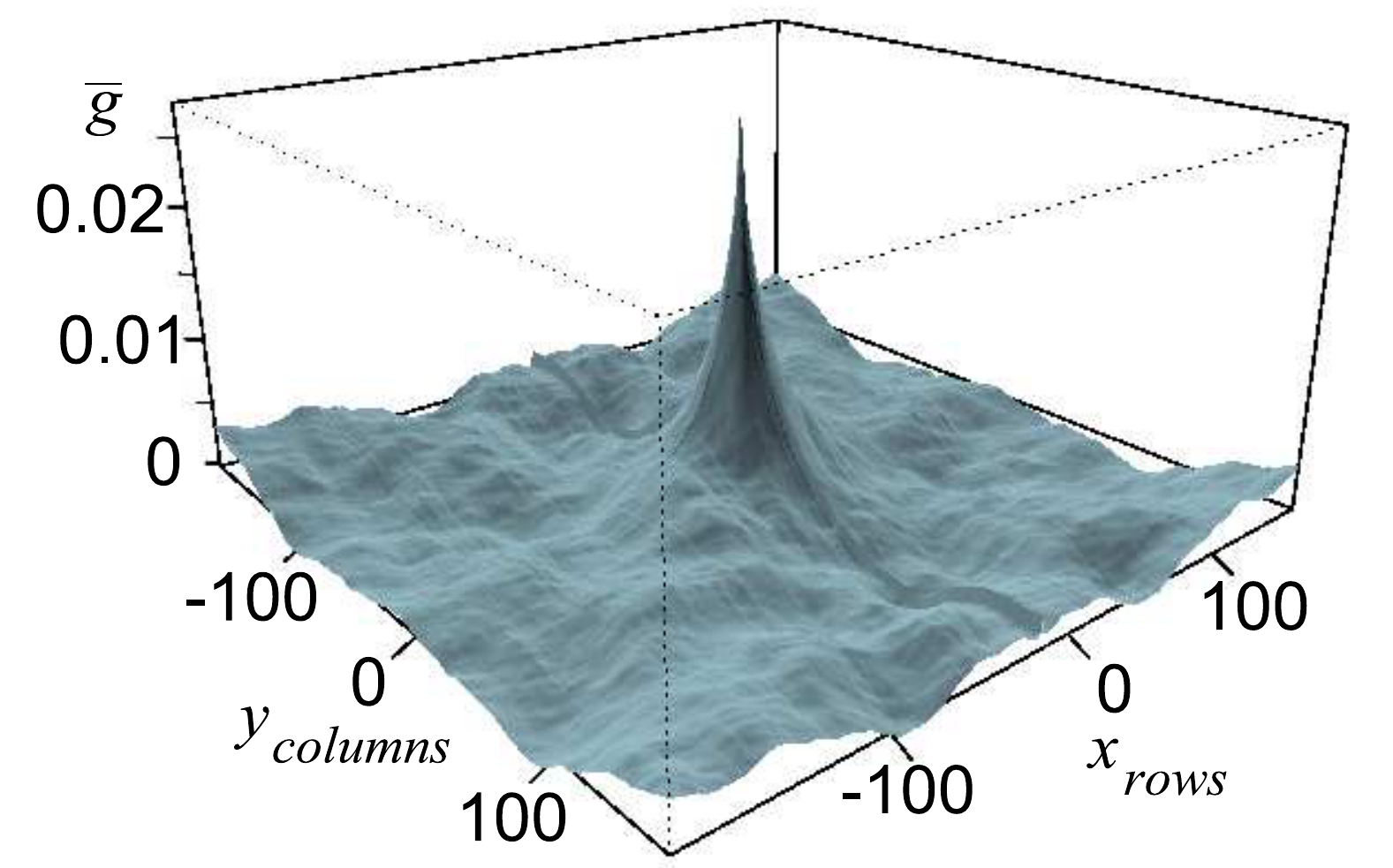} &
\includegraphics[width=0.5\columnwidth]{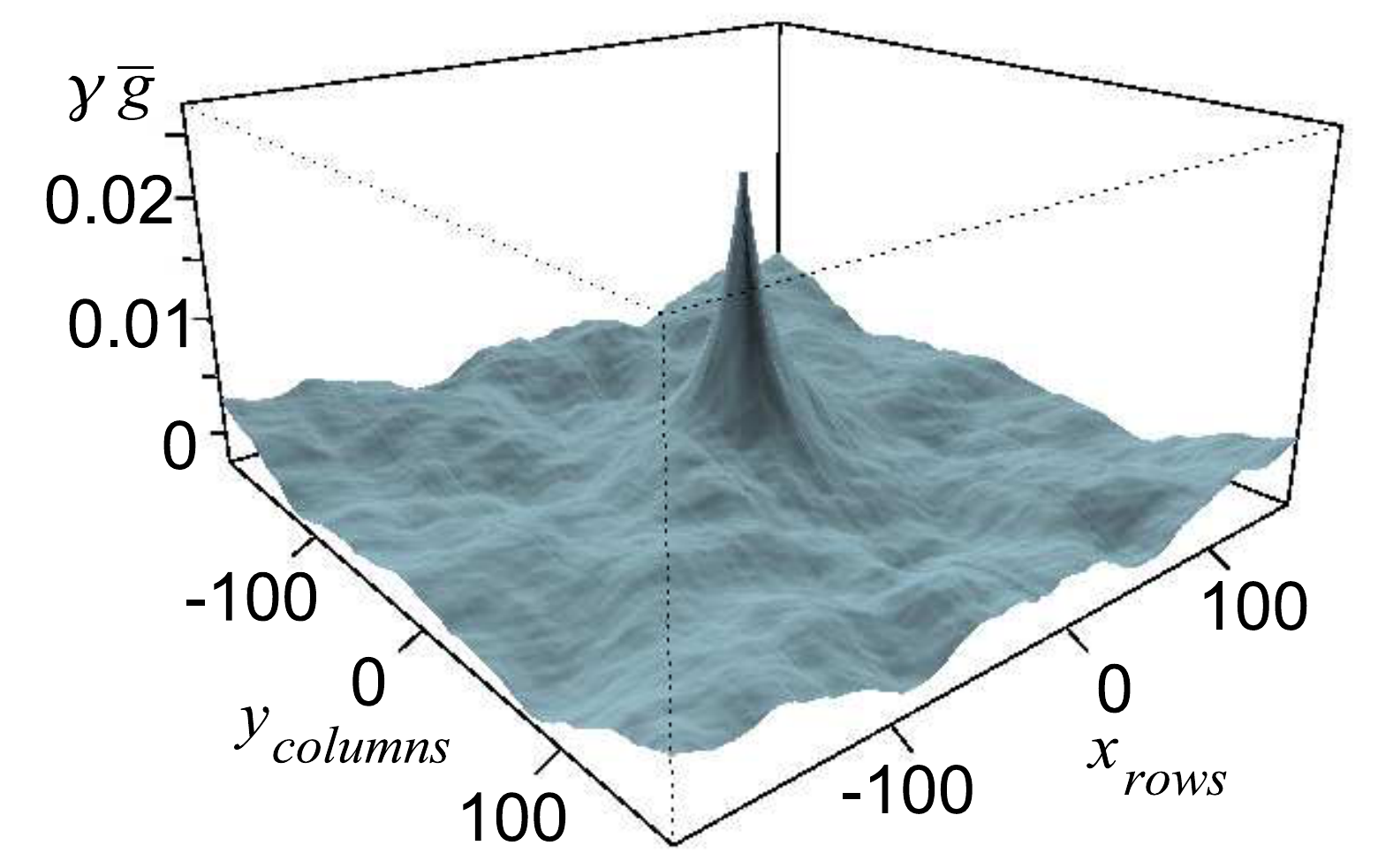}\\
{\small $\overline{g}_{\alpha_{1}}(h)$} & {\small $\gamma\overline{g}%
_{\alpha_{1}}(h)$}\\
\includegraphics[width=0.5\columnwidth]{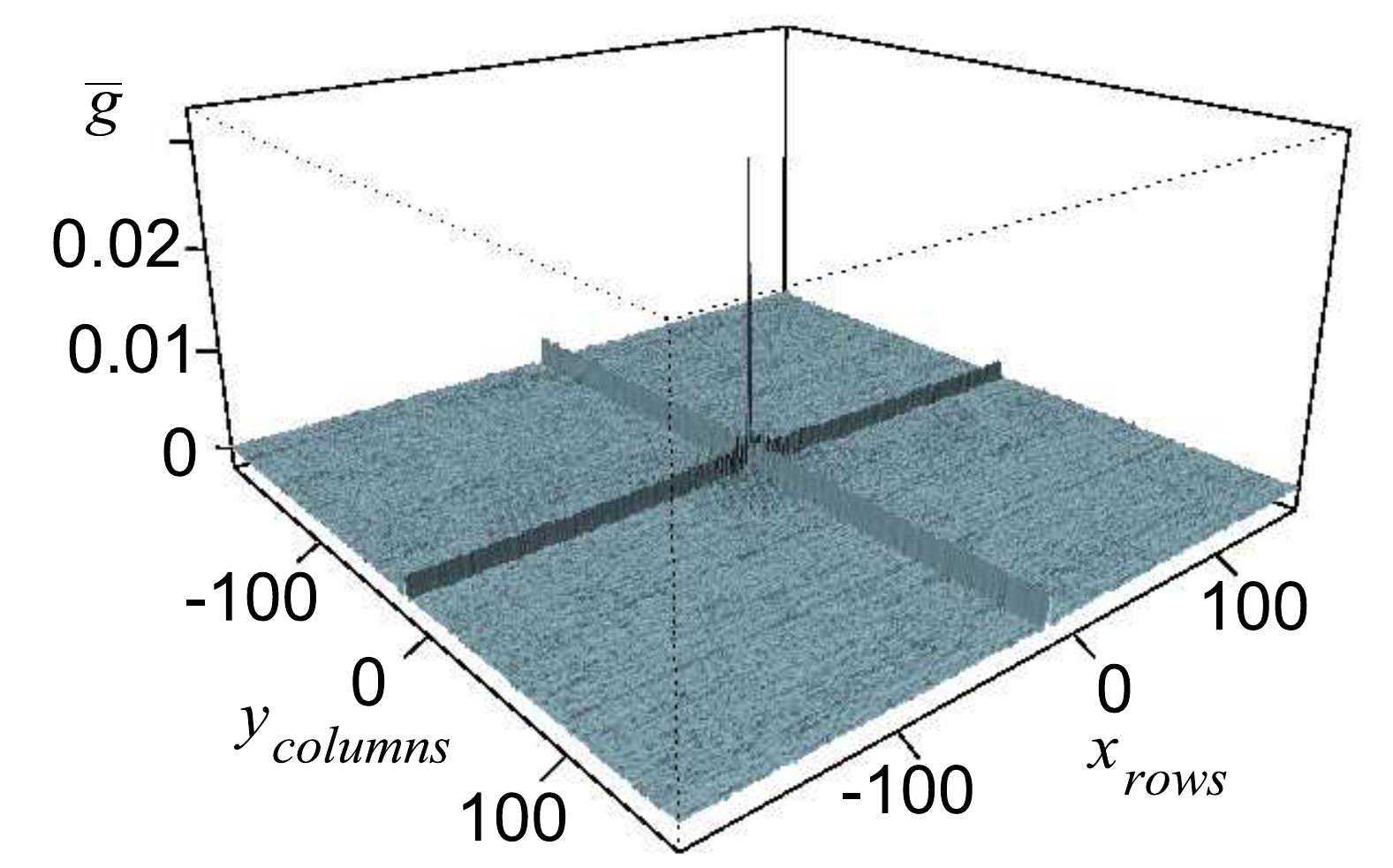} &
\includegraphics[width=0.5\columnwidth]{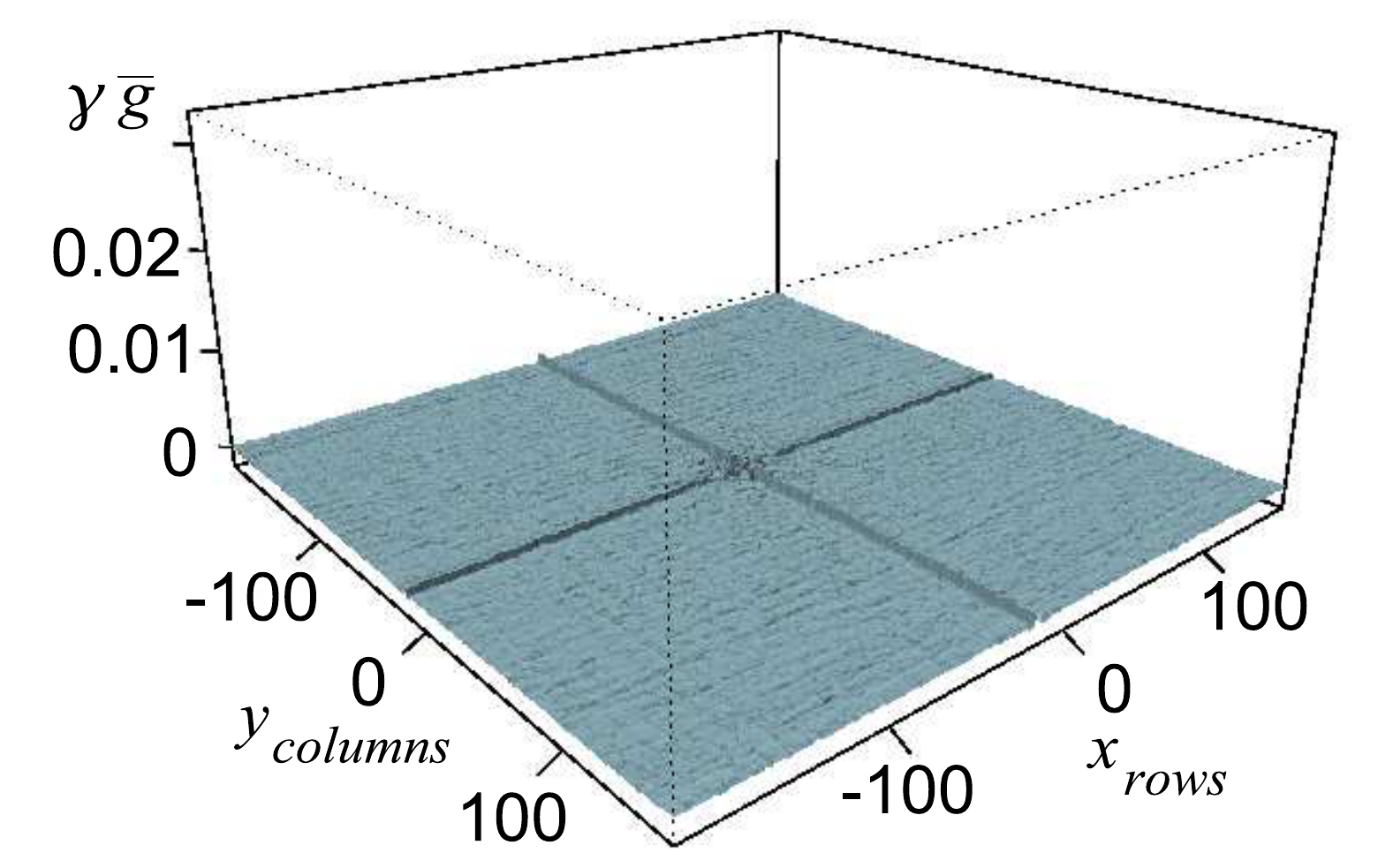}\\
{\small $\overline{g}_{\alpha_{2}}(h)$} & {\small $\gamma\overline{g}%
_{\alpha_{2}}(h)$}\\
&
\end{tabular}
\caption{2D covariances for factor pixels
$c^{\mathbf{f}}_{\alpha_{1}}$ and $c^{\mathbf{f}}_{\alpha_{2}}$ of
image ``Roujan" before (left) and after
opening (right).}%
\label{fig_covariance}%
\end{figure}

The covariance peak, at the origin, contains the sum of the signal
variance and the noise variance of the image. Then, the signal
variance is estimated by the maximum (i.e. value at the origin) of
the covariance $\overline{g}$. Based on the property of the
morphological opening which removes intensity peaks smaller than the
used structuring element, we apply a morphological opening $\gamma$
on the covariance image with a unitary centered structuring element
(square of $3\times3$ pixels) in order to remove the peak of signal
associated to the noise variance:
$Var(signal)=\gamma\overline{g}_{\alpha _{k}}(0)$. The noise
variance is therefore given by the residue of the opening of the
covariance at the
origin: $Var(noise)=\overline{g}_{\alpha_{k}}(0)-\gamma\overline{g}%
_{\alpha_{k}}(0)$ (fig. \ref{fig_covariance}). Hence, the signal to
noise ratio is defined for a factor axis $\alpha_{k}$ as:
\begin{equation}
SNR_{\alpha_{k}}=\frac{Var(signal)}{Var(noise)}=\frac{\gamma\overline
{g}_{\alpha_{k}}(0)}{\overline{g}_{\alpha_{k}}(0)-\gamma\overline{g}%
_{\alpha_{k}}(0)}\label{eq_SNR}%
\end{equation}

In the current example, by analysis of the factor pixels and their signal to
noise ratios, we observe that they are higher for axes 1, $c_{\alpha_{1}%
}^{\mathbf{f}}$, and 3, $c_{\alpha_{3}}^{\mathbf{f}}$, than for axes
2, $c_{\alpha_{2}}^{\mathbf{f}}$, and 4,
$c_{\alpha_{4}}^{\mathbf{f}}$. Therefore, the axes 1 and 3 are
retained. For an automatic selection of factor axes, the axes with a
SNR lower than 1 are rejected.

\begin{figure}
\centering
\begin{tabular}
[c]{@{}c@{ }c@{}}%
\includegraphics[width=0.49\columnwidth]{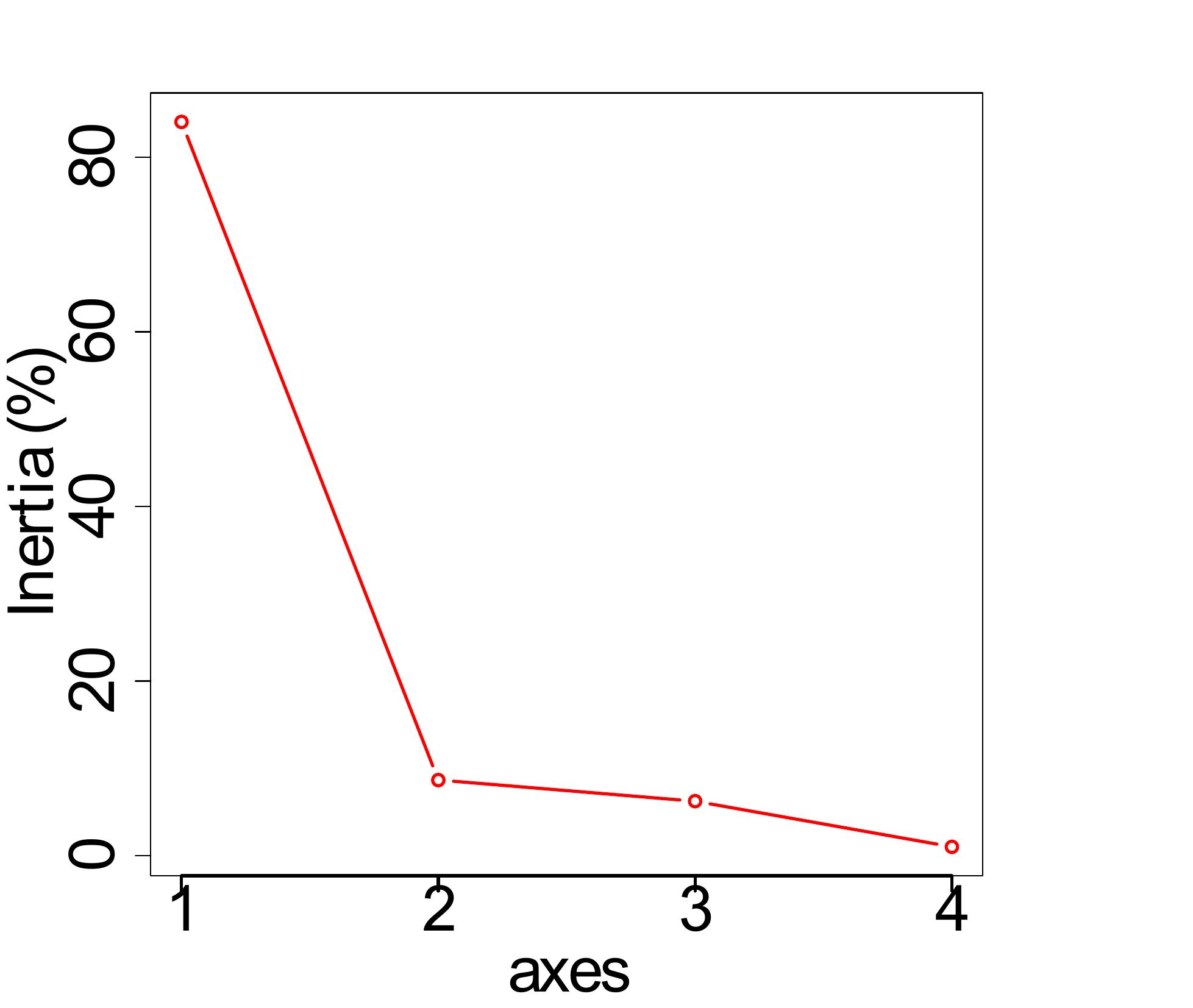} &
\includegraphics[width=0.49\columnwidth]{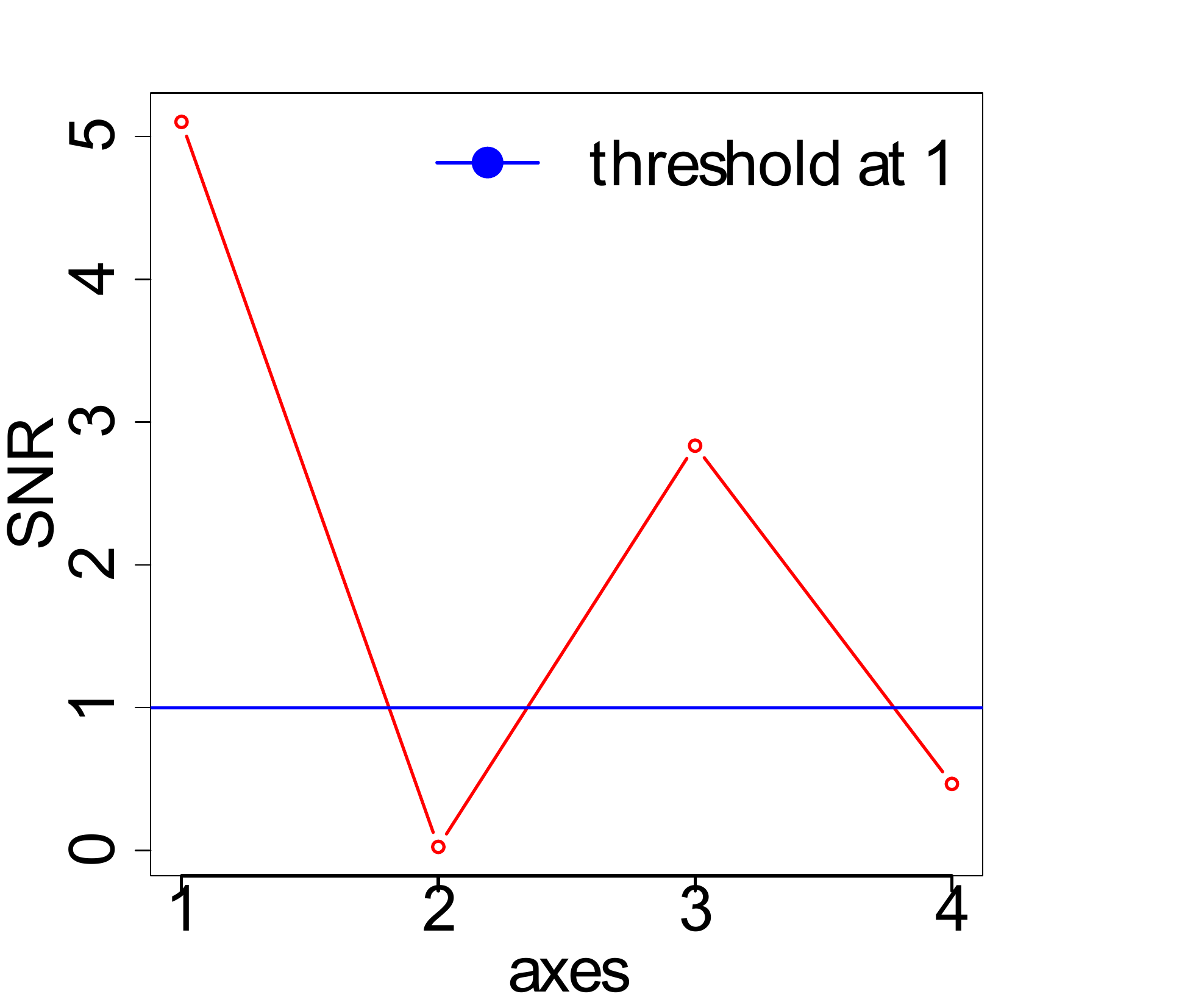}\\
&
\end{tabular}
\caption{Inertia (\%) and SNR for factor pixels of image ``Roujan".}%
\label{fig_SNR_inertia}%
\end{figure}

In figure \ref{fig_SNR_inertia}, we notice that axis 3, which is selected, has
a lower inertia than axis 2, which is rejected. Therefore, SNR analysis makes
it possible to describe relevant signal in dimensionality reduction, more than
inertia-like criterion.

\FloatBarrier

\section{Introduction to stochastic WS}
\label{sec_intro_sto_WS}

\citet{AnguloJeulin_ISMM_2007} defined a new method of stochastic WS
for greyscale and color images. This method was  extended to
hyperspectral images by \citet{Noyel_KES_2007}. In addition, the
segmentation of multispectral images by classical (deterministic) WS
was presented in \citep{Noyel_IAS_2007}. Several improvements were
made by \citet{Noyel_CGIV_2008,Noyel_PhD_2008}.

\subsection{Principle of the stochastic WS}

One of the main artefacts of the classical watershed is that small
regions strongly depend on the position of the markers, or on the
volume (i.e. the integral of the grey levels) of the catchment
basin, associated to their minima. In fact, there are two kinds of
contours associated to the watershed of a gradient: first order
contours, which correspond to significant regions and which are
relatively independent from markers; second order contours,
associated to ``small", ``low" contrasted and textured regions,
which depend strongly on the location of markers. Stochastic
watershed aims at enhancing the first order contours from a sampling
effect, to improve the result of the watershed segmentation.

Let us consider $\{mrk_{i}(x)\}_{i=1}^{M}$ a series of $M$
realisations of $N$ uniform or regionalized random germs. Each of
these binary images is considered as the marker for a watershed
segmentation of a scalar gradient or a vector gradient.
Therefore, a series of segmentations is obtained, i.e. $\{sg_{i}%
^{mrk}(x)\}_{i=1}^{M}$. Starting from the $M$ realisations of
contours, the probability density function $pdf(x)$ of contours is
computed by the Parzen window method. The kernel density estimation
by Parzen window \citep{Duda_1973} is a way to estimate the
probability density function (pdf) of a random
variable. Let $\mathbf{x}_{1},\mathbf{x}_{2},\cdots,\mathbf{x}_{M}%
\in\mathbb{R}^{n}$ be $M$ samples of a random variable, the kernel density
approximation of its pdf is: $\widehat{f}_{h}(\mathbf{x})=\frac{1}{Nh}%
\sum_{i=1}^{N}K\left(  \frac{\mathbf{x}-\mathbf{x}_{i}}{h}\right) $,
where $K(\mathbf{x})$ is some kernel and the bandwidth $h$ a
smoothing parameter. Usually, $K(\mathbf{x})$ is taken to be a
Gaussian function, $G_{\sigma}$, with mean zero and variance
$\sigma^{2}$. The smoothing effect of the Gaussian convolution
kernel (typically $\sigma=3$ working on contours of one pixel width)
is important to obtain a function where near contours, such as
textured regions or associated to small regions, are added together.
In other words, the WS lines with a very low probability, which
correspond to non significant boundaries, are filtered out.

The pdf of contours could be thresholded to obtain the most prominent
contours. However, to obtain closed contours, the pdf image can be also
segmented, as we consider here, using a watershed segmentation.

\FloatBarrier
\subsection{Influence of parameters}

The stochastic WS needs two parameters:

\begin{enumerate}

\item $M$ realisations of germs. The method is almost independent on $M$ if it
is large enough. Practically, the convergence is ensured for $M$ in
the range 20~-~50. We propose to use $M$ equal to 100.

\item $N$ germs (or markers): In WS segmentation the number of
regions obtained is equal to the number of markers. In the case of
the stochastic WS, if $N$ is small, a segmentation in large regions
is privileged; if $N$ is too large, the over-segmentation of
$sg_{i}^{mrk}$ leads to a very smooth $pdf$, which looses its
properties to select the $R$ regions. Therefore, the stochastic WS
mainly depends on $N$, but is linked to the number of regions $R$
which should be finally selected. As it was shown in
\citep{AnguloJeulin_ISMM_2007}, it is straightforward to use $N>R$.
\end{enumerate}

As it is described below, in section \ref{sec_seg_sto_WS}, in this
study the number of markers of each realization used in the
conditioned stochastic WS will be equal to the number of connected
classes of the classification $\widehat{\kappa}$ different of the
void class.

\FloatBarrier
\section{Generation of markers from a classification}
\label{sec_generation_markers_classification}

To segment the pdf of contours by WS, it is necessary to have
markers for the spectral objects of interests. The latter are
obtained by a spectral classification. This classification step
groups the pixels into classes with similar spectra. In fact, each
pixel is compared to all the others, leading to a global comparison
of the image spectra. We have tested two kinds of methods for
spectral classification on multispectral images: unsupervised
methods and supervised methods.

In the case of remote sensing, we show results obtained by an
unsupervised method such as ``clara" \citep{Kaufman_1990}.
Unsupervised and supervised methods in remote sensing using
mathematical morphology as well as the ``clara" method are also
investigated in \citep{Epifanio_2007}. For other contexts different
from remote sensing, we tested other unsupervised methods such as
k-means and supervised methods such as Linear Discriminant Analysis
\citep{Noyel_ISBI_2008,Noyel_PhD_2008}. One of the strong points of
supervised approaches is to integrate prior information about
spectrum classes into the classification. However, in generic remote
sensing segmentation problems, we are not looking for a special kind
of spectrum. Therefore, we prefer, in the presentation of the
current paper, to use an unsupervised classification with a unique
parameter, namely the number of classes.

\begin{figure}
\centering
\begin{tabular}
[c]{@{}c@{ }c@{ }c@{}}%
\includegraphics[width=0.33\columnwidth]{roujan_Synthetic_RGB_R=chan3_G=chan2_B=chan1_bis.png} &
\includegraphics[width=0.33\columnwidth]{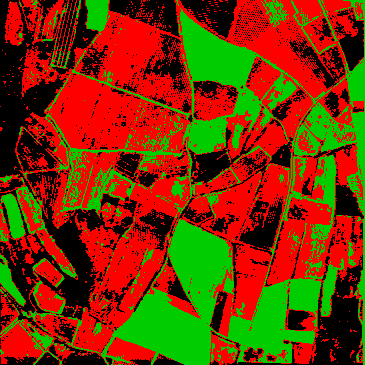} &
\includegraphics[width=0.33\columnwidth]{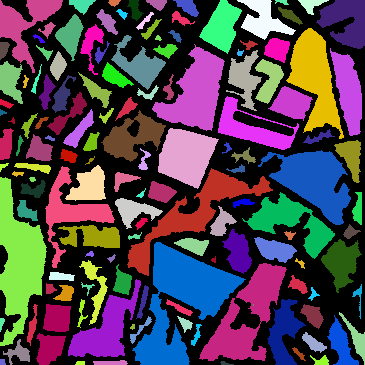}\\
{\small \emph{(a)} RGB ``Roujan"} & {\small \emph{(b)} $\kappa$} & {\small \emph{(c)} $\widehat{\kappa}$}\\
\includegraphics[width=0.33\columnwidth]{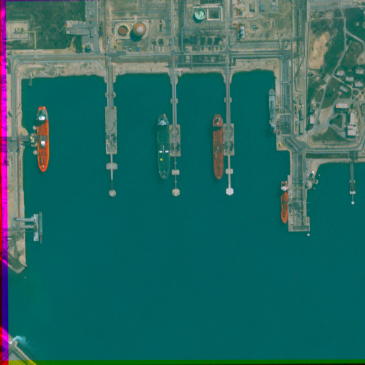} &
\includegraphics[width=0.33\columnwidth]{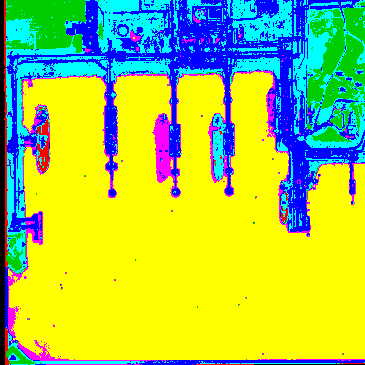} &
\includegraphics[width=0.33\columnwidth]{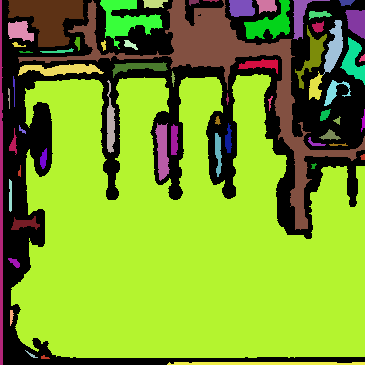}\\
{\small \emph{(d)} RGB ``Port de Bouc"} & {\small \emph{(e)} $\kappa$} & {\small \emph{(f)} $\widehat{\kappa}$%
}\\
&  &
\end{tabular}
\caption{First line: \emph{(a)} Synthetic RGB image ``Roujan",
\emph{(b)} classification $\kappa$ in 3 classes in factor space
formed by $c^{\mathbf{f}}_{\alpha_{1}}$ and
$c^{\mathbf{f}}_{\alpha_{3}}$ and \emph{(c)} transformed
classification $\widehat{\kappa }=\Upsilon(\kappa)$. Second line:
\emph{(d)} Synthetic RGB image ``Port de Bouc", \emph{(e)}
classification $\kappa$ in 7 classes in factor space formed by $c^{\mathbf{f}%
}_{\alpha_{1}}$ and $c^{\mathbf{f}}_{\alpha_{2}}$ and \emph{(f)}
transformed classification $\widehat{\kappa}=\Upsilon(\kappa)$. The
void class is in black on the images $\widehat{\kappa}$. The way to
compute the transformed classification is explained in section
\ref{sec_transfomed_classification}. The colours correspond to the
labels of the
transformed classification obtained after a new labellisation.}%
\label{Fig_classification_roujan}%
\end{figure}

Hence, after the data reduction and the spectral filtering stage by
FCA, a spectral classification by ``clara" is performed on the
factor space. We have to stress that a correct classification, used
later for constructing the pdf, requires factor pixels without
noise. The classification algorithm uses an Euclidean distance which
is coherent with the metric of the factor space
\citep{Noyel_IAS_2007}. The unique parameter is the number of
classes, which is chosen in order to get a good separation of \ the
classes. Here, for this example, 3 classes are retained (fig.
\ref{Fig_classification_roujan}).

\FloatBarrier
\section{Multispectral segmentation by stochastic watershed}
\label{sec_seg_sto_WS}

The previously obtained spectral classification of pixels leads to a
spatial partition of the image on regions which are spectrally
homogeneous. This initial partition is denoted $\kappa$. However,
they do not define a partition into spatial connected classes:
various connected components belong to the same spectral class. In
order to address this issue, a segmentation stage is needed which
introduces spatial information (i.e., regional information). In this
section, after presenting an algorithm to pre-process the previous
classification $\kappa$, the extension of stochastic WS to
multispectral images is explained; then we show a way to constrain
the pdf by the previous spectral classification.

\FloatBarrier
\subsection{Pre-processing of markers coming from \ the classification}
\label{sec_transfomed_classification}

A pre-processing of the connected classes of the classification is necessary
for two reasons:

\begin{enumerate}
\item to give  necessary degrees of freedom to the final WS of the pdf.
Actually, if we use the connected classes of the classification
$\kappa$ as markers of the flooding process, then the WS is
completely defined by the limit of the classes after labelling them
into connected classes. Therefore we propose to process the initial
partition using an anti-extensive transformation such as an erosion
\citep{Serra_1982,Soille_1999}.  For instance, using a structuring
element (SE) of size 5 $\times$ 5 pixels each spatial class is
reduced and the smallest classes disappear. We consider that classes
corresponding to ``noise'' are totally removed. The smallest classes
could also be removed by means of an area opening
\citep{Soille_1999}. Therefore, after processing the initial
partition, a partial partition is obtained. There are several
alternatives to deal with the problem that the partial partition is
not a partition of the support space; for instance adding to the
partial partition the singletons outside its support. We prefer
however the following solution. We introduce a particular label, the
``void class'', in such a way that after processing the initial
partition, covering the whole space, a new partition is also
obtained. In the new partition, the existing classes are modified
(reduced) and in addition, a new class is introduced, the void
class. If we want to keep all the classes, for instance if the aim
is to segment thin objects as a network of roads, we could use a
homotopic thinning instead of an erosion.
\item to fill the holes inside the largest classes with an extensive
transformation such as a closing by reconstruction with a SE of size
for instance 3 $\times$ 3 pixels \citep{Serra_1982,Soille_1999}.
\end{enumerate}

The transformations are applied to each class of the initial
spectral classification. More precisely to each spectral class,
$\kappa(n)$ with $n=1,\ldots,N$, of the previous classification
$\kappa=\cup_{n}\kappa(n)$, an index function (i.e. a binary image)
is associated:
\begin{equation}
h_{\kappa(n)}(x)=\left\{
\begin{array}
[c]{lll}%
1, & \text{if} & x\in\kappa(n);\\
0, & \text{otherwise.} & \\
\end{array}
\right.  \label{eq_index_function}%
\end{equation}
Then, the sequence of the anti-extensive and extensive
transformations are applied to each index function, $h_{\kappa(n)}$
for all $n=1,\ldots,N$, in order to obtain the transformed index
function $h_{\kappa(n)}^{\prime}$. Then the supremum (union) of the
index functions is computed, followed by a standard labelling to
obtain the connected components which correspond to markers for the
segmentation, except for the void class which is not a marker. The
complete transform is noted $\Upsilon$, and it results in a new
partition $\widehat{\kappa}=\Upsilon(\kappa)$ (fig.
\ref{Fig_classification_roujan}). The partition $\widehat{\kappa}$
will then be strongly used for the spatial segmentation by
stochastic WS.

\FloatBarrier

\subsection{Extension of the stochastic WS to multivariate or hyperspectral
images}

The extension of the stochastic WS was first presented in
\citet{Noyel_KES_2007}. The key points are recalled in the current
paper.

\subsubsection{Spectral distances and gradient}

In order to segment images according to watershed-based paradigms, a
gradient is needed. A gradient image, in fact its norm, is a scalar
function with values in the reduced interval $[0,1]$ (after
normalisation), i.e. $\varrho(x):E\rightarrow \lbrack0,1]$. In order
to define a gradient, two approaches are considered: the standard
symmetric morphological gradient on each marginal channel and a
metric-based vectorial gradient on all channels.

The morphological gradient is defined for scalar images $f$ as the
difference between dilation and erosion by a unit structuring
element $B$, i.e.,
\begin{eqnarray*}
  \nonumber \varrho(f_{\lambda_j}(x)) &=& \delta_B(f_{\lambda_j}(x))-\varepsilon_B(f_{\lambda_j}(x)) \\
  \nonumber &=& \vee[ f_{\lambda_j}(y), y \in B(x) ] -
    \wedge[ f_{\lambda_j}(y), y \in B(x) ]\text{ .}
\end{eqnarray*}
As it was shown in \citet{Hanbury_2001}, the morphological gradient
can be generalised to vectorial image by using the following
relation:
\begin{equation*}
   \{f(x) - \vee[f(y), y \in B(x), y \neq x]\} = \wedge\{f(x) - f(y), y \in B(x), y \neq x\}
\end{equation*}
and the relation obtained by the inversion of the suprema and
infima. The morphological gradient can be written in a form that is
only composed of increments computed in the neighborhood $B$
centered at point $x$:
\begin{equation*}
    \varrho(f)(x) = \vee[f(x) - f(y), y \in B(x), y \neq x] - \wedge[f(x) - f(y), y \in B(x), y \neq x]\text{ .}
\end{equation*}
To extend this relation to multivariate functions, it is sufficient
to replace the increment $[f(x) - f(y)]$ by a distance between
vector pixels $d(\mathbf{f_{\lambda}}(x), \mathbf{f_{\lambda}}(y))$
to obtain the following metric-based gradient:
\begin{equation*}
    \varrho_{d}\mathbf{f_{\lambda}}(x) = \vee[ d(\mathbf{f_{\lambda}}(x),
    \mathbf{f_{\lambda}}(y)) , y \in B(x), y \neq x] -
    \wedge[ d(\mathbf{f_{\lambda}}(x), \mathbf{f_{\lambda}}(y)) , y \in
    B(x), y \neq x
] \text{ .}
\end{equation*}

Various metric distances, useful for multispectral images, are
available for this gradient such as:
\begin{itemize}
 \item the Euclidean distance:
\[
d_{E}(\mathbf{f_{\lambda}}(x), \mathbf{f_{\lambda}}(y)) = \sqrt{
\sum_{j=1}^{L}( f_{\lambda_{j}}(x) - f_{\lambda_{j}}(y) )^2 },
\]
and
\item the Chi-squared distance:
\[
d_{\chi^{2}}( \mathbf{f}_{\lambda}(x_{i}) ,
\mathbf{f}_{\lambda}(x_{i'}) ) = \sqrt{\sum_{j=1}^{L}
\frac{S}{f_{.\lambda_{j}}} \left( \frac{ f_{\lambda_{j}}(x_{i}) }{
f_{x_{i}.} } - \frac{ f_{\lambda_{j}}(x_{i'}) }{ f_{x_{i'}.} }
\right)^{2}}
\]
with $f_{.\lambda_{j}} = \sum_{i=1}^{P}
f_{\lambda_{j}}(x_{i})$, $f_{x_{i}.} = \sum_{j=1}^{L}
f_{\lambda_{j}}(x_{i})$ and $S =
\sum_{j=1}^{L}\sum_{i=1}^{P}f_{\lambda_{j}}(x_{i})$.
\end{itemize}
An important point is to choose an appropriate distance depending on
the space used for image representation: Chi-squared distance is
adapted to MIS and Euclidean distance to FIS. More details on
multivariate gradients are given by \citet{Noyel_IAS_2007}. Another
example of a multivariate gradient is given by
\citet{Scheunders_2002}.

\FloatBarrier
\subsubsection{Probability density function for multispectral images}

\begin{figure}
\begin{center}
\includegraphics[width=0.9\columnwidth]{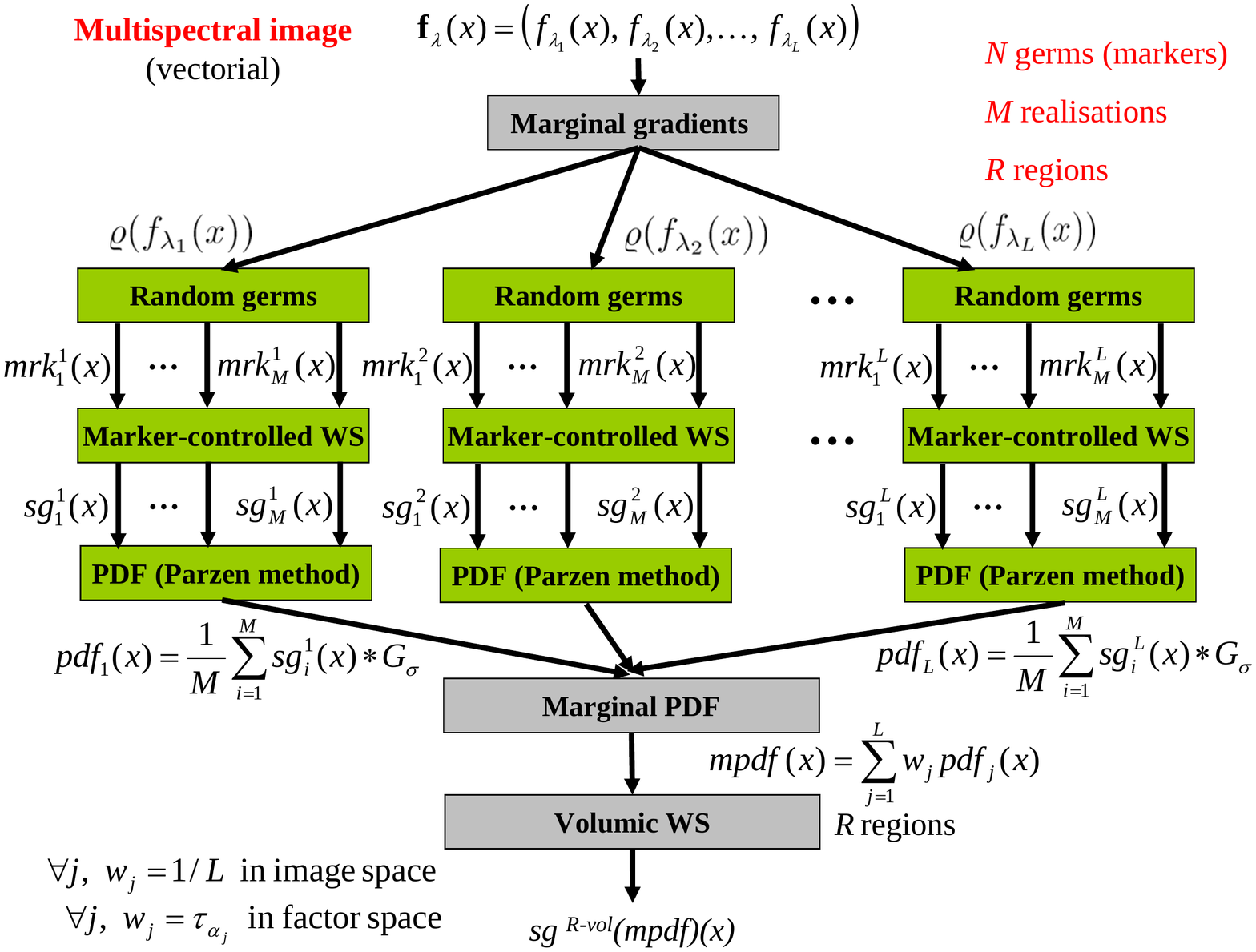}\newline
\end{center}
\caption{General framework stochastic WS on marginal pdf $mpdf$ for
multispectral images}%
\label{fig_sto_WS_mpdf_framework}%
\end{figure}

\begin{figure}
\begin{center}
\includegraphics[width=0.9\columnwidth]{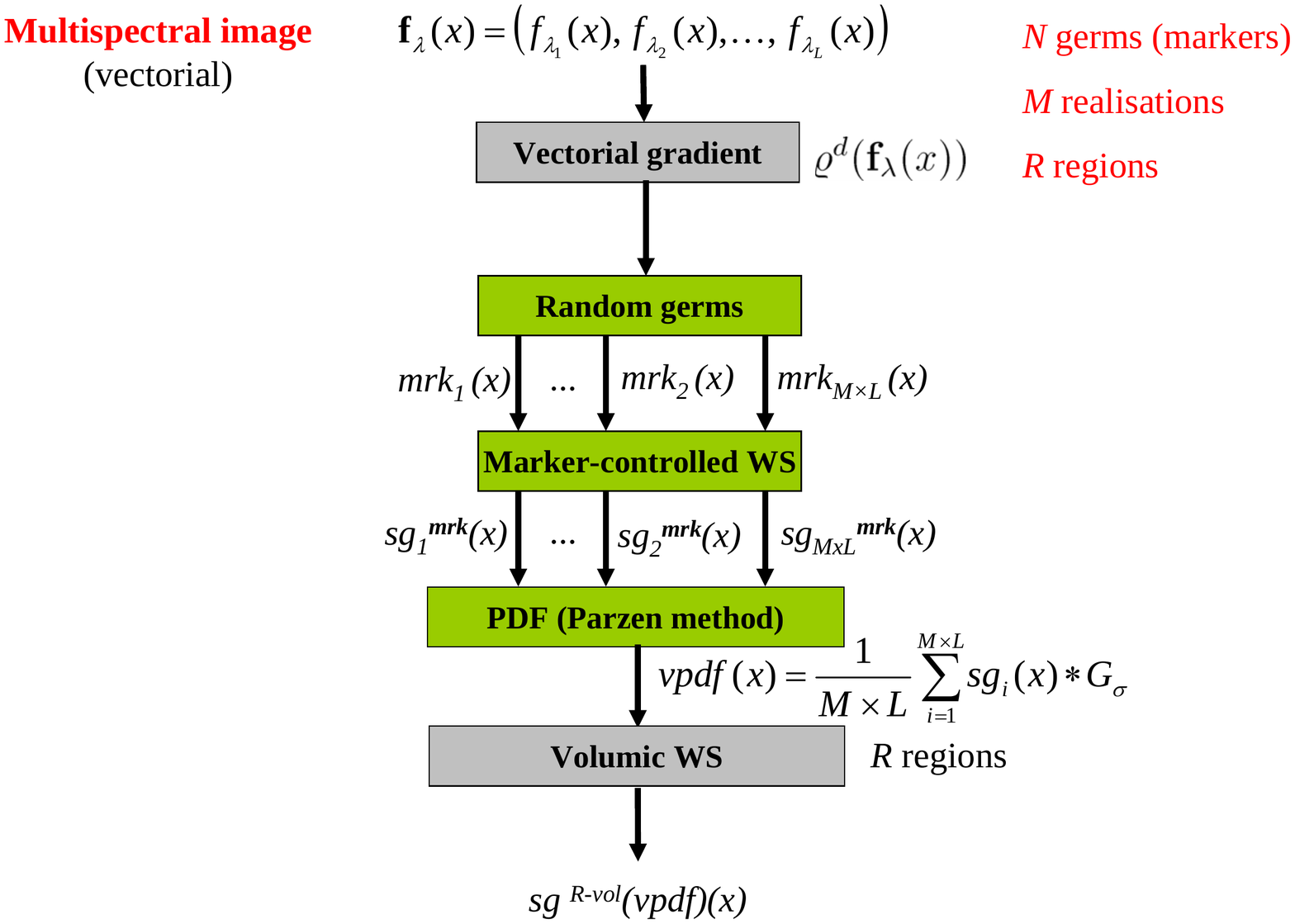}\newline
\end{center}
\caption{General framework stochastic WS on vectorial pdf $vpdf$ for
multispectral images}%
\label{fig_sto_WS_vpdf_framework}%
\end{figure}

We studied two ways to extend the pdf \ gradient to multispectral
images:

\begin{enumerate}

\item the first one is a marginal approach (i.e. channel by  channel) called
marginal pdf $mpdf$ (alg. \ref{alg_mpdf} and fig.
\ref{fig_sto_WS_mpdf_framework})

\item the second one is a vectorial approach (i.e. vector pixel by vector
pixel) called vectorial pdf $vpdf$ (alg. \ref{alg_vpdf} and fig.
\ref{fig_sto_WS_vpdf_framework}).
\end{enumerate}

\begin{algorithm}[!htb]
\caption{$mpdf$} \label{alg_mpdf}
%\algsetup{indent = 2em}
\begin{algorithmic}[1]
\STATE For the morphological gradient of each channel
$\varrho(f_{\lambda_{j}})$, $j \in [1, \ldots, L]$, throw $M$
realisations of $N$ uniform random germs, i.e. the markers
$\{mrk^j_i\}_{i=1 \ldots M}^{j=1 \ldots L}$, generating $M \times L$
realisations. Get the series of segmentations,
$\{sg^{j}_{i}(x)\}_{i=1 \ldots M}^{j=1 \ldots L}$, by watershed
associated to morphological gradients of each channel
$\varrho(f_{\lambda_{j}})$. \STATE  Get the marginal pdfs on each
channel by Parzen method: $pdf_j(x) =  \frac{1}{M} \sum_{i=1}^M
sg^j_i(x) \ast G_{\sigma}$. \STATE Obtain the weighted marginal pdf:
\begin{equation}\label{eq_mpdf}
mpdf(x) = \sum_{j=1}^L w_j pdf_j (x)
\end{equation}
with $w_j=1/L$, $j \in [1, \ldots, L]$ in MIS and $w_j$ equal to
the inertia axes in FIS.
\end{algorithmic}
\end{algorithm}

\begin{algorithm}[!htb]
\caption{$vpdf$} \label{alg_vpdf}
\algsetup{indent = 2em}
\begin{algorithmic}[1]
\STATE For the vectorial gradient $\varrho^{d}(\mathbf{f_{\lambda}})$,
throw $M\times L$ realisations of $N$ uniform random germs, i.e. the markers
$\{mrk_i\}_{i=1 \ldots M \times L}$, with $L$ the channels number. Get the segmentation,
$\{sg_{i}(x)\}_{i=1 \ldots M \times L}$, by watershed associated to the vectorial gradient
$\varrho^{d}(\mathbf{f_{\lambda}})$, with $d=d_{\chi^2}$ in MIS or $d=d_{E}$ in FIS.
\STATE Obtain the probability density function:
\begin{equation}\label{eq_vpdf}
vpdf(x) = \frac{1}{M\times L} \sum_{i=1}^{M \times L} sg_i(x) \ast
G_{\sigma} \text{ .}
\end{equation}
\end{algorithmic}
\end{algorithm}

The probabilistic gradient was also defined in
\citep{AnguloJeulin_ISMM_2007} to ponder the enhancement of the
largest regions by the introduction of smallest regions. It is
defined as $\varrho_{prob} = mpdf + \varrho^{d}$: after
normalization in $[0,1]$ of the weighted marginal pdf $mpdf$ and the
metric-based gradient $\varrho^{d}$.

\begin{figure}
\centering
\begin{tabular}
[c]{cccc}%
\includegraphics[width=0.22\columnwidth]{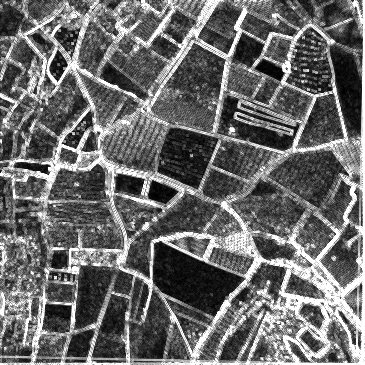} &
\includegraphics[width=0.22\columnwidth]{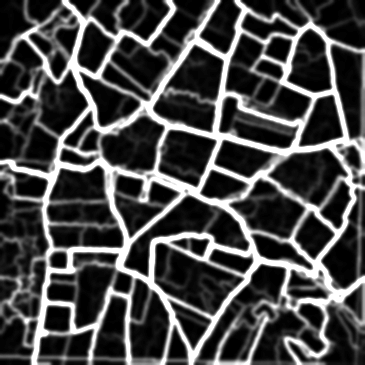} &
\includegraphics[width=0.22\columnwidth]{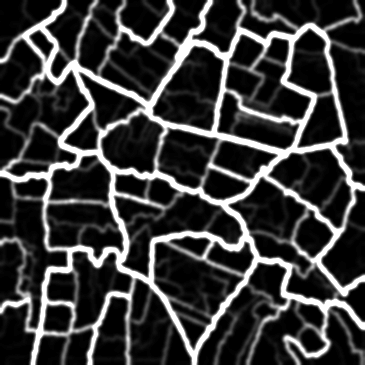} &
\includegraphics[width=0.22\columnwidth]{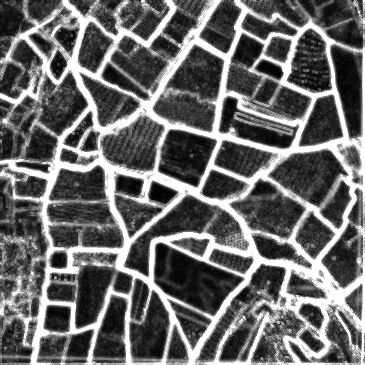}\\
{\small \emph{(a)} $\varrho^{\chi^{2}}(\mathbf{f_{\lambda}})$} &
{\small \emph{(b)} $mpdf(\mathbf{f_{\lambda}})$} & {\small
\emph{(c)} $vpdf(\mathbf{f_{\lambda}})$}
& {\small \emph{(d)} $\varrho_{prob}(\mathbf{f_{\lambda}})$}\\
\includegraphics[width=0.22\columnwidth]{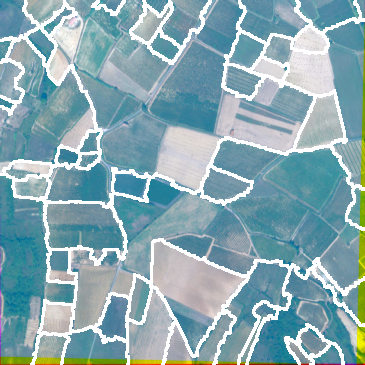} &
\includegraphics[width=0.22\columnwidth]{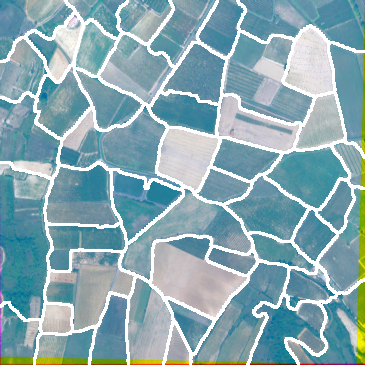} &
\includegraphics[width=0.22\columnwidth]{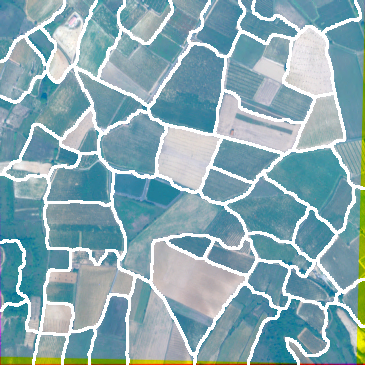} &
\includegraphics[width=0.22\columnwidth]{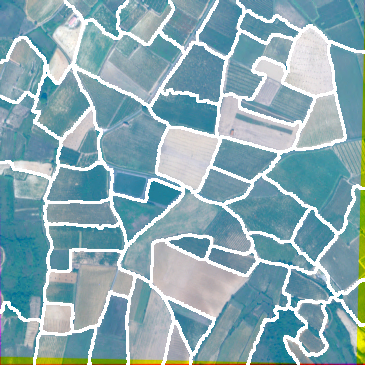}\\
{\small \emph{(e)}
$seg^{vol}(\varrho^{\chi^{2}}(\mathbf{f}_{\lambda}),R)$} &
{\small \emph{(f)} $seg^{vol}(mpdf(\mathbf{f}_{\lambda}),R)$} & {\small \emph{(g)} $seg^{vol}%
(vpdf(\mathbf{f}_{\lambda}),R)$} & {\small \emph{(h)}
$seg^{vol}(\varrho_{prob}(\mathbf{f}_{\lambda}),R)$}\\
&  &  &
\end{tabular}
\caption{Top: \emph{(a)} Gradients and \emph{(b, c, d)} pdf; bottom:
\emph{(e, f, g, h)} associated WS segmentations, with a volume
criterion, on image ``Roujan", in MIS, with $N = 50$ points , $M =
100$
realisations, $R=50$ regions with the largest volume.}%
\label{fig_segmentation_roujan}%
\end{figure}

In order to obtain a partition from the $mpdf$, $vpdf$ or the
gradient $\varrho_{prob}$, these probabilistic functions can be
segmented for instance by a hierarchical WS with a volume criterion
(see the examples of fig. \ref{fig_segmentation_roujan}), as studied
in \citet{Noyel_KES_2007}. In such a case, the goal is not to find
all the regions. In fact, the stochastic WS addresses the problem of
segmentation of an image in few pertinent regions, according to a
combined criterion of contrast and size. In the present study, as we
discuss below, the segmentation of the pdf is obtained from markers
of the processed classification. In \citet{Noyel_KES_2007} was also
shown that marginal pdf and vectorial pdf give similar results in
multispectral image space (MIS) and in factor image space (FIS), as
we can also observe in comparison of fig.
\ref{fig_segmentation_roujan}. Consequently, in what follows, we
choose to work on the MIS to show that our method is not limited by
the number of channels.

\FloatBarrier

\subsection{Conditioning the germs of the pdf by a previous classification}

The pdf of contours with uniform random germs is constructed without
any prior information about the spatial/spectral distribution of the
image. In this part, we introduce spectral information by
conditioning the germs by the previous transformed classification
$\widehat{\kappa}$. Therefore this obtained pdf contains
spatio-spectral information. It is possible to use point germs or
random ball germs whose location is conditioned by the
classification. An exhaustive study is presented by
\citet{Noyel_PhD_2008,Noyel_ECMI_2008}. In the sequel we present
random ball germs regionalized by a classification where each
connected class may be hit one time, $mrk_{i}^{\kappa-b}(x)$.

Let us explain the procedure. The transformed classification
$\widehat{\kappa}$ is composed of connected classes,
$\widehat{\kappa}=\cup_{k}C_{k}$ with $C_k \cap C_{k'} = \emptyset$,
for $k \ne k'$. The void class is written $C_{0}$. Then the random
germs are drawn conditionally to the connected components $C_{k}$ of
the filtered classification $\widehat{\kappa}$. To do this, the
following rejection method is used: the random point germs are
uniformly distributed. If a point germ $m$ falls inside a connected
component $C_{k}$ of minimal area $S$ and not yet marked, then it is
kept, otherwise it is rejected. Therefore not all the germs are
kept. These point germs are called random point germs regionalized
by the classification $\kappa$. However, the regionalized point
germs are sampling all the classes independently of their prior
estimate of class size/shape, given by the classification. In order
to tackle with this limitation, we propose to use random balls as
germs.

The centers of the balls are the random point germs and the radii
$r$ are uniformly distributed between $0$ and a maximum radius
$Rmax$: $\mathcal{U}[1,Rmax]$. Only the intersection between the
ball $B(m,r)$ and the connected component $C_{p}$ is kept as a germ.
These balls are called random balls germs regionalized by the
classification $\kappa$ and noted $mrk_{i}^{\kappa-b}(x)$.

The algorithm \ref{alg_random_balls_germs_regionalized} sketches the process.
We notice that $N$ is the number of random germs to be generated. The
effective number of implanted germs is less than $N$.

\begin{algorithm}[!htb]
\caption{Random ball germs regionalized by a classification (each
connected class may be hit one time) $mrk_i^{\kappa-b}(x)$}
\label{alg_random_balls_germs_regionalized} \algsetup{indent = 2em}
\begin{algorithmic}[1]
\STATE Given $N$ the number of drawn germs $m$
\STATE Set the background class and the void class $C_0$ to \emph{marked}
\FORALL{ drawn germs $m$ from 1 to $N$ }
\IF{$C_k$, such as $m \in C_k$, is \emph{not marked}}
\STATE $r = \mathcal{U}[1,Rmax]$
\STATE Keep as a germ  $\mathrm{B}(m,r) \cap C_k$
\STATE Set the class $C_k$ to \emph{marked}
\ENDIF
\ENDFOR
\end{algorithmic}
\end{algorithm}

\begin{figure}
\centering
\begin{tabular}
[c]{@{}c@{ }c@{ }c@{ }c@{ }c@{}}%
{\small i = 1} & {\small i = 2} & {\small i = 3} & {\small i = 4} &
{\small i
= 5}\\
\includegraphics[width=0.2\columnwidth]{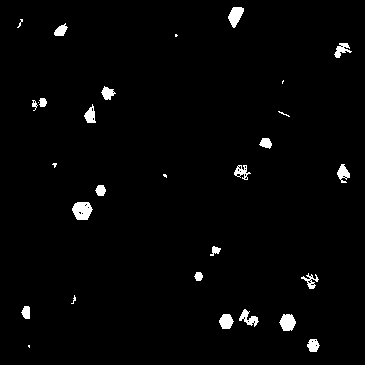} &
\includegraphics[width=0.2\columnwidth]{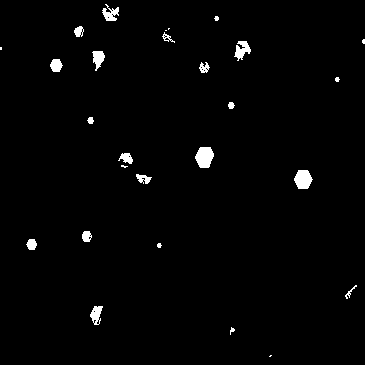} &
\includegraphics[width=0.2\columnwidth]{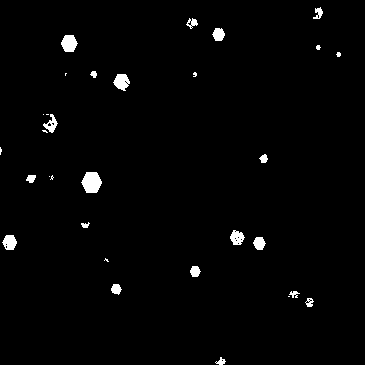} &
\includegraphics[width=0.2\columnwidth]{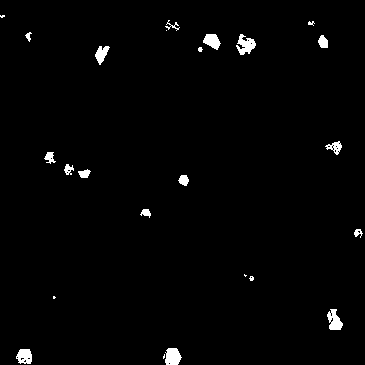} &
\includegraphics[width=0.2\columnwidth]{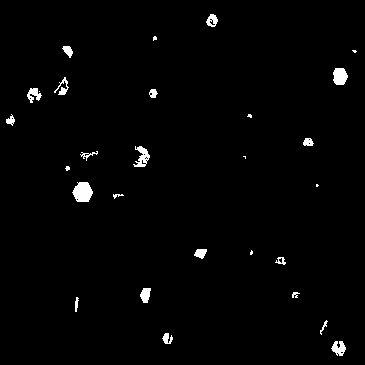}\\
{\small \emph{(a)} $mrk_{1}^{\kappa-b,1}(x)$} & {\small \emph{(b)}
$mrk_{2}^{\kappa-b,1}(x)$} & {\small \emph{(c)}
$mrk_{3}^{\kappa-b,1}(x)$} & {\small \emph{(d)}
$mrk_{4}^{\kappa-b,1}(x)$} &
{\small \emph{(e)} $mrk_{5}^{\kappa-b,1}(x)$}\\
\includegraphics[width=0.2\columnwidth]{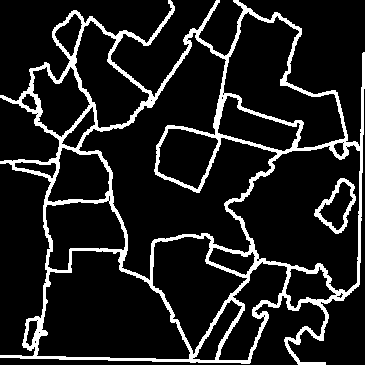} &
\includegraphics[width=0.2\columnwidth]{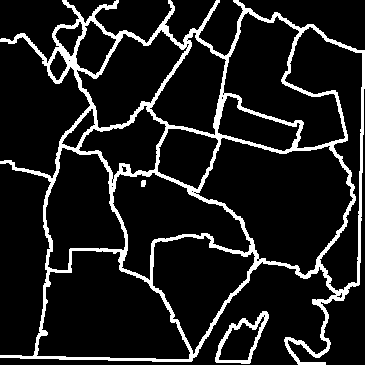} &
\includegraphics[width=0.2\columnwidth]{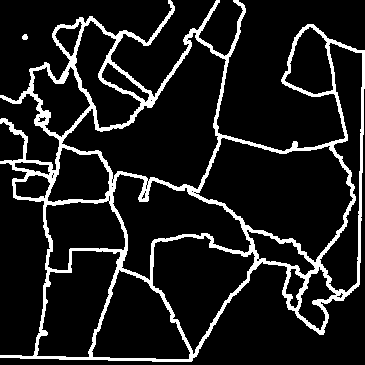} &
\includegraphics[width=0.2\columnwidth]{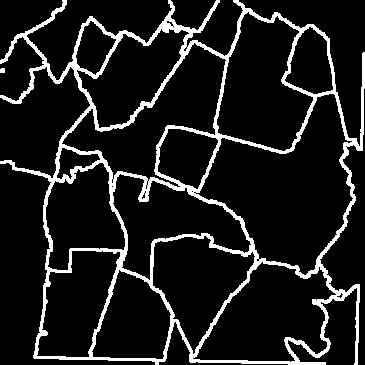} &
\includegraphics[width=0.2\columnwidth]{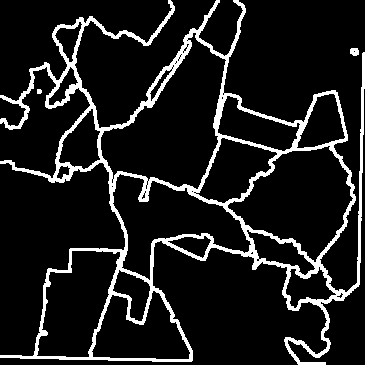}\\
{\small \emph{(f)} $sg_{1}^{1}(x)$} & {\small \emph{(g)}
$sg_{2}^{1}(x)$} & {\small \emph{(h)} $sg_{3}^{1}(x)$}
& {\small \emph{(i)} $sg_{4}^{1}(x)$} & {\small \emph{(j)} $sg_{5}^{1}(x)$}\\
&  &
\includegraphics[width=0.2\columnwidth]{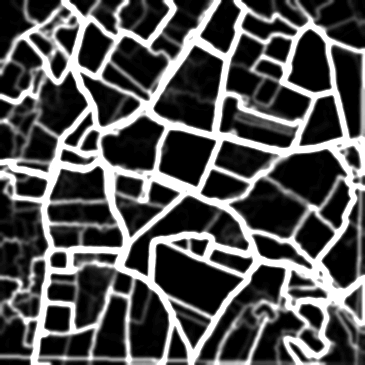} &
& \\
&  & {\small \emph{(k)} $mpdf^{\kappa-b}(\mathbf{f}_{\lambda})$} &  & \\
&  &  &  &
\end{tabular}
\caption{\emph{(a, b, c, d, e)} A few realisations of regionalized random ball germs $\{mrk_{i}%
^{\kappa-b,j}(x) \}_{i=1 \ldots5}^{j=1}$ (top) by the classification
$\kappa$ and \emph{(f, g, h, i, j)} associated contours of watershed
$\{sg_{i}^{j}(x) \}_{i=1 \ldots5}^{j=1}$ (bottom). $i$ is the index
of realisations and $j$ the number of the considered channel. The
parameters are the maximum radius $Rmax$ = 30 pixels and the maximum
number of germs $N$ = 50. \emph{(k)} The marginal pdf with
regionalized random ball germs
$mpdf^{\kappa-b}(\mathbf{f}_{\lambda})$ is computed with
$M=100$ realisations.}%
\label{Fig_realisations_mpdf_roujan}%
\end{figure}

We use the marginal pdf to show our results. Some random ball germs
regionalized by a classification, $\{mrk_{i}^{\kappa-b,j}(x) \}_{i=1
\ldots 5}^{j=1}$, their associated realisations of contours,
$\{sg_{i}^{j}(x) \}_{i=1 \ldots5}^{j=1}$, and the marginal pdf
computed in MIS, $mpdf^{\kappa -b}(\mathbf{f}_{\lambda})$, are
presented in figure \ref{Fig_realisations_mpdf_roujan}.

\begin{figure}
\centering
\begin{tabular}
[c]{@{}c@{ }c@{}}%
\includegraphics[width=0.33\columnwidth]{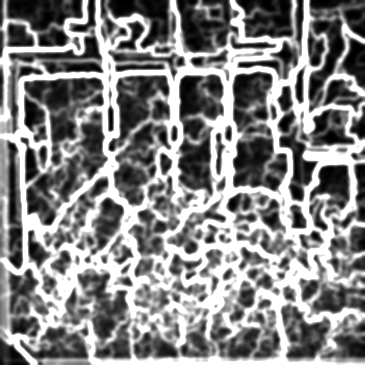} &
\includegraphics[width=0.33\columnwidth]{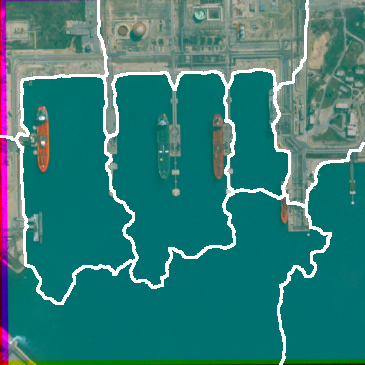}\\
{\small \emph{(a)} { $mpdf^{pt}(\mathbf{f}_{\lambda})$ }} & {\small \emph{(b)} { $sg^{R-vol}%
(mpdf^{pt}(\mathbf{f}_{\lambda}))$ }}\\
\includegraphics[width=0.33\columnwidth]{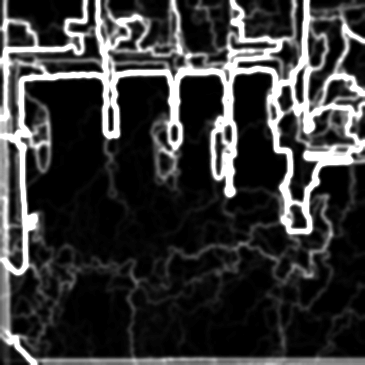} &
\includegraphics[width=0.33\columnwidth]{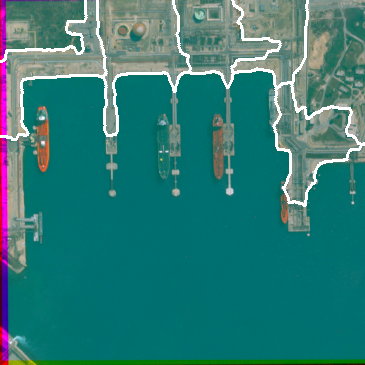}\\
{\small \emph{(c)} { $mpdf^{\kappa-b}(\mathbf{f}_{\lambda})$ }} &
{\small \emph{(d)} {
$sg^{R-vol}(mpdf^{\kappa-b}(\mathbf{f}_{\lambda}))$ }}\\
&
\end{tabular}
\caption{Comparison between \emph{(a)} the pdf
$mpdf^{pt}(\mathbf{f}_{\lambda})$ and \emph{(b)} the associated
stochastic WS with uniform random point germs with a volume
criterion $sg^{R-vol}(mpdf^{pt}(\mathbf{f}_{\lambda}))$ to
\emph{(c)} the pdf $mpdf^{\kappa-b}(\mathbf{f}_{\lambda})$ and
\emph{(d)} the associated stochastic WS
with regionalized random ball germs with a volume criterion $sg^{R-vol}%
(mpdf^{\kappa-b}(\mathbf{f}_{\lambda}))$. Only $R=8$ regions with
the largest volume are retained. For visualization, the contours of
the segmentations are
dilated by a SE of size $3 \times3$ pixels.}%
\label{Fig_comparaison_port_de_bouc_WSstoballs_WSsto}%
\end{figure}

A comparison between the marginal pdf with uniform random point
germs, $mpdf^{pt}(\mathbf{f}_{\lambda})$ (noted before
$mpdf(\mathbf{f}_{\lambda})$)
and the marginal pdf with random ball germs $mpdf^{\kappa-b}(\mathbf{f}%
_{\lambda})$ is shown in figure
\ref{Fig_comparaison_port_de_bouc_WSstoballs_WSsto}. These pdf are segmented
by hierarchical WS with a volume criterion, $sg^{R-vol}(mpdf^{pt}%
(\mathbf{f}_{\lambda}))$ and
$sg^{R-vol}(mpdf^{\kappa-b}(\mathbf{f}_{\lambda }))$. In this
particular example, the advantage of the pdf obtained with the
conditioned random balls is obvious with respect to the pdf from
uniform random point germs. Moreover, in order to compare the
results of segmentation of example of figure
\ref{Fig_comparaison_port_de_bouc_WSstoballs_WSsto}, we must notice
that in both images only the $8$ most important regions have been
segmented. If the purpose is to segment all the objects, including
the smallest ones, we only need to increase the number of desired
volumic regions.

\FloatBarrier

\subsection{Segmentation of the pdf}

\label{part_seg_pdf}

After having built the pdf, a segmentation stage is required to obtain the
contours of the image. Two possibilities were tested:

\begin{enumerate}

\item a marker-controlled WS using as markers the transformed  classification
$sg^{mrk}(mpdf^{\kappa-b}(\mathbf{f}_{\lambda}) , \widehat{\kappa})$
\citep{Noyel_CGIV_2008}. In this  case, no more parameter is required.

\item a hierarchical WS with a volume criterion  $sg^{R-vol}(mpdf^{\kappa
-b}(\mathbf{f}_{\lambda}))$  \citep{Noyel_KES_2007}. Therefore, it is
necessary to define an additional parameter: the  number of regions with the
largest volume.
\end{enumerate}

In order to show the robustness of our approach we used the same
method to build the pdf conditioned by the classification on similar
images ``Roujan", ``Roujan 0 2" and ``Roujan 0 9". As for image
``Roujan", factor axes $c^{\mathbf{f}}_{\alpha_{1}}$ and
$c^{\mathbf{f}}_{\alpha_{3}}$ are kept because th Signal to Noise
Ratio (SNR) is less than thresholded value (1). Inertias and SNR for
factor axes of images ``Roujan X X" are presented in the tables
\ref{tab_Inertia_axes_roujan} and \ref{tab_SNR_axes_roujan}.

\begin{table}
\tbl{Part of inertia for factor axes of images ``Roujan X X"} {%
\begin{tabular}
[c]{c|cccc}%
Image & $c^{\mathbf{f}}_{\alpha_{1}}$ &
$c^{\mathbf{f}}_{\alpha_{2}}$ & $c^{\mathbf{f}}_{\alpha_{3}}$ &
$c^{\mathbf{f}}_{\alpha_{4}}$\\\hline
``Roujan" & 84.1 \% & 8.7 \% & 6.2 \% & 1 \%\\
``Roujan 0 2" & 75.6 \% & 13.7 \% & 9.2 \% & 1.5 \%\\
``Roujan 1 9" & 77.5 \% & 12.1 \% & 9 \% & 1.4 \%\\
\end{tabular}
} \label{tab_Inertia_axes_roujan}\end{table}

\begin{table}
\tbl{SNR for factor axes of images ``Roujan X X"} {%
\begin{tabular}
[c]{c|cccc}%
Image & $c^{\mathbf{f}}_{\alpha_{1}}$ &
$c^{\mathbf{f}}_{\alpha_{2}}$ & $c^{\mathbf{f}}_{\alpha_{3}}$ &
$c^{\mathbf{f}}_{\alpha_{4}}$\\\hline
``Roujan" & 5.10 & 0.03 & 2.84 & 0.47\\
``Roujan 0 2" & 2.66 & 0.01 & 2.53 & 0.32\\
``Roujan 1 9" & 3.15 & 0 & 2.90 & 0.51\\
\end{tabular}
} \label{tab_SNR_axes_roujan}\end{table}

A classification ``clara" is made on the factor pixels of these two
axes. The marginal pdf, $mpdf^{\kappa-b}(\mathbf{f}_{\lambda})$, is
produced as explained for image ``Roujan".

The parameters for the stochastic WS controlled by markers $sg^{mrk}%
(mpdf^{\kappa-b}(\mathbf{f}_{\lambda}) , \widehat{\kappa})$ are:

\begin{itemize}
\item the number of classes for the classification $\kappa$:  $Q=3$. It is the
only important parameter;

\item the maximum number of random ball germs: $N=50$, which must be high
enough and which must be of the the same order as the number of
regions in the image;

\item the number of realisations for each channel: $M=100$, which always has
the same value;

\item the minimum area $S=10$ pixels for the connected classes of the
classification, generally the same;

\item the parameters for pre-processing the initial spectral
classification are an erosion using a $5 \times 5$ square and a
closing by reconstruction using as marker a dilation using a $3
\times 3$ square, generally the same;

\item the maximum radius of the random ball $Rmax=30$, generally the same.
\end{itemize}

\begin{figure}
\centering
\begin{tabular}{cccc}
    \begin{tabular}{c}
    \small{~Roujan~}
    \end{tabular}&
    \includegraphics[width=0.25\columnwidth]{roujan_marqueurs_classif_fond_noir.png}&
    \includegraphics[width=0.25\columnwidth]{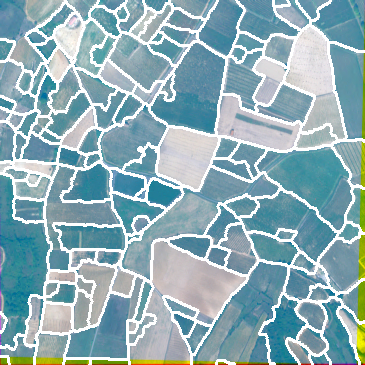}&
    \includegraphics[width=0.25\columnwidth]{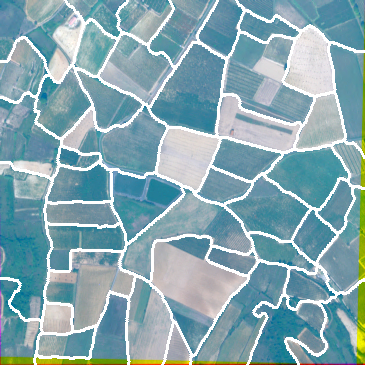}\\
    &
    \small{ \emph{(a)} $\widehat{\kappa}$ }&
    \small{ \emph{(b)} $sg^{mrk}(mpdf^{\kappa-b}(\f) , \widehat{\kappa})$ }&
    \small{ \emph{(c)} $sg^{R-vol}(mpdf^{\kappa-b}(\f))$ }\\
    && \small{ 140 regions } & \small{ 50 regions } \\
    \hline\\
    \begin{tabular}{c}
    \small{~Roujan 0 2~}
    \end{tabular}&
    \includegraphics[width=0.25\columnwidth]{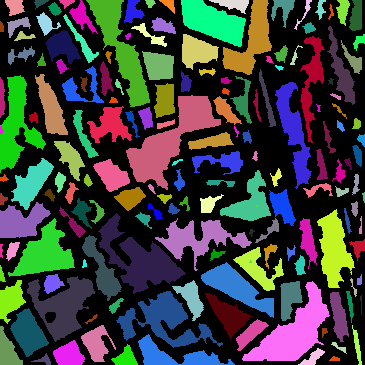}&
    \includegraphics[width=0.25\columnwidth]{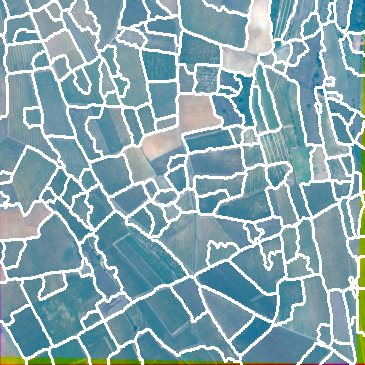}&
    \includegraphics[width=0.25\columnwidth]{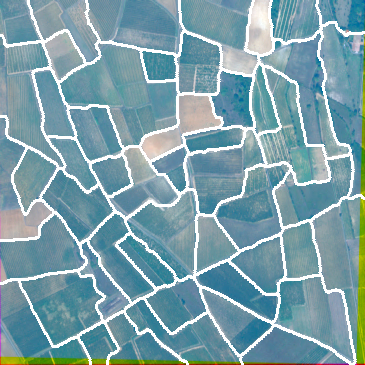}\\
    &
    \small{ \emph{(d)} $\widehat{\kappa}$ }&
    \small{ \emph{(e)} $sg^{mrk}(mpdf^{\kappa-b}(\f) , \widehat{\kappa})$ }&
    \small{ \emph{(f)} $sg^{R-vol}(mpdf^{\kappa-b}(\f))$ }\\
    && \small{ 162 regions } & \small{ 50 regions } \\
    \hline\\
    \begin{tabular}{c}
    \small{~Roujan 1 9~}
    \end{tabular}&
    \includegraphics[width=0.25\columnwidth]{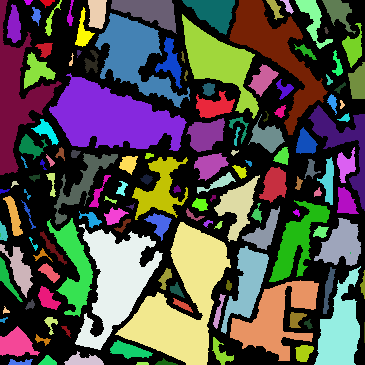}&
    \includegraphics[width=0.25\columnwidth]{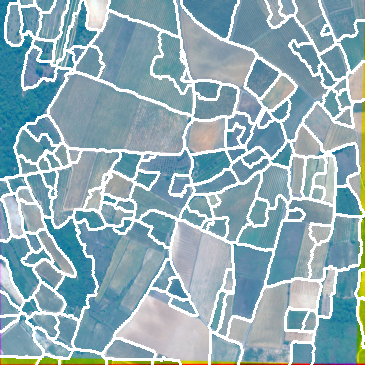}&
    \includegraphics[width=0.25\columnwidth]{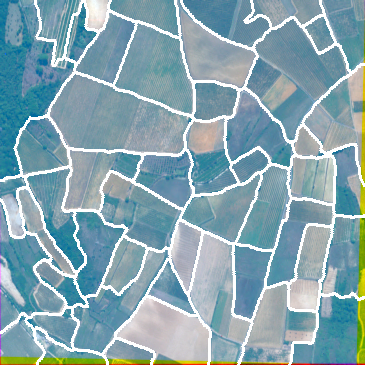}\\
    &
    \small{ \emph{(g)} $\widehat{\kappa}$ }&
    \small{ \emph{(h)} $sg^{mrk}(mpdf^{\kappa-b}(\f) , \widehat{\kappa})$ }&
    \small{ \emph{(i)} $sg^{R-vol}(mpdf^{\kappa-b}(\f))$ }\\
    && \small{ 142 regions } & \small{ 50 regions } \\
\end{tabular}
  \caption{Comparison between  segmentations $sg^{mrk}(mpdf^{\kappa-b}(\f) , \widehat{\kappa})$ \emph{(b, e, h)} with markers
   coming from the filtered classification $\widehat{\kappa}$ \emph{(a, d, g)}
    and segmentations $sg^{R-vol}(mpdf^{\kappa-b}(\f))$ \emph{(c, f, i)} by stochastic WS with a prior given number of
    regions $R = 50$.
   The results are given for images: ``Roujan",
  ``Roujan 0 2", ``Roujan 1 9".}
  \label{Fig_segmentations_WS_marqueurs_ou_volumique}
\end{figure}

The final segmentations with marker-controlled WS,
$sg^{mrk}(mpdf^{\kappa -b}(\mathbf{f}_{\lambda}) ,
\widehat{\kappa})$, and hierarchical WS with a volume criterion,
$sg^{R-vol}(mpdf^{\kappa-b}(\mathbf{f}_{\lambda}))$, are shown on
figure \ref{Fig_segmentations_WS_marqueurs_ou_volumique}. We notice
that the number of regions, resulting from the segmentation,
strongly depends on the image, whereas the number of spectral
classes is the same for similar images. In fact, the number of
regions for volumic WS, $sg^{vol}$, must be chosen according to the
spatial diversity of the considered image, and cannot be easily
fixed ``a priori" (50 here). The number of regions depends on the
size and the complexity of the image, while the number of spectral
classes depends on the spectral content, which for a specific domain
can be relatively constant. Consequently, it is more relevant to
segment the pdf with markers coming also from the classification
$sg^{mrk}$. Indeed, only one parameter is needed for
marker-controlled WS $sg^{mrk}(mpdf^{\kappa-b}(\mathbf{f}_{\lambda})
, \widehat{\kappa})$: the number of classes $Q$ in the
classification $\kappa$. Moreover, in that sense, the approach based
on the number of classes $Q$ produces a more robust segmentation
than the approach based on the number of regions $R$.

In order to show that the stochastic WS, conditioned by a previous
classification, produces contours which are more regular and robust
than deterministic (standard) WS, we compare the results of  a
deterministic approach to a stochastic one:

\begin{enumerate}

\item For both cases, an unsupervised classification $\kappa$ (``clara") is
processed in factor space FIS and transformed $\widehat{\kappa} =
\Upsilon(\kappa)$.

\item
\begin{itemize}

\item For the deterministic approach, a chi-squared metric based  gradient
$\varrho^{\chi^{2}}(\mathbf{f}_{\lambda})$  is computed in image
space MIS, as a function to flood.

\item For the stochastic approach, a marginal probability density function
$mpdf^{\kappa-b}(\mathbf{f}_{\lambda})$, with regionalized random
ball germs conditioned by the classification $\kappa$,  is
processed, in image space MIS, as a function to flood.
\end{itemize}

\item In both cases, the flooding function is segmented by a  watershed (WS)
using as sources of flooding the markers from the transformed
classification $\widehat{\kappa}$.
\end{enumerate}

\begin{figure}
\centering
\begin{tabular}
[c]{c}%
\begin{tabular}
[c]{cc}%
\includegraphics[width=0.25\columnwidth]{roujan_classif_clara_3classes_axesAFC_1_3.png} &
\includegraphics[width=0.25\columnwidth]{roujan_marqueurs_classif_fond_noir.png}\\
\small{\emph{(a)} $\kappa$} &
\small{\emph{(b)} $\widehat{\kappa}$}\\
&
\end{tabular}
\\%
\begin{tabular}
[c]{ccc}%
\includegraphics[width=0.25\columnwidth]{gradVectorial_Chi2_roujan_bis.png} &
\includegraphics[width=0.25\columnwidth]{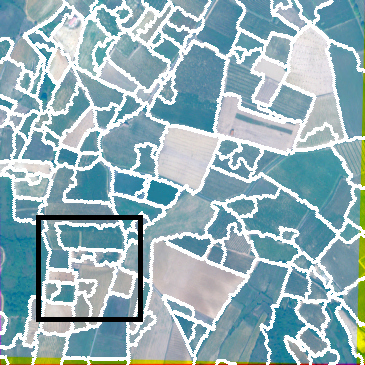} &
\includegraphics[width=0.25\columnwidth]{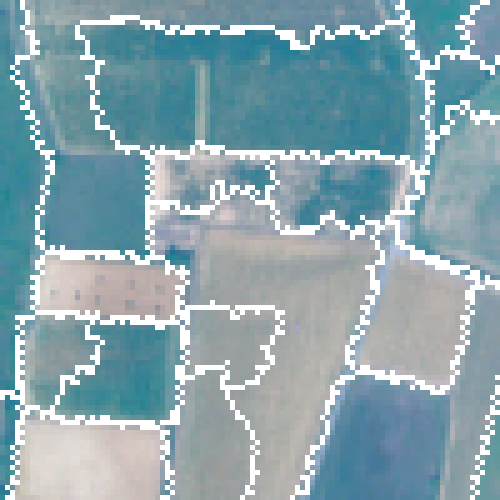}\\
\small{\emph{(c)} $\varrho^{\chi^{2}}(\mathbf{f}_{\lambda})$} &
\small{\emph{(d)}
$sg^{mrk}(\varrho^{\chi^{2}}(\mathbf{f}_{\lambda}),\widehat{\kappa})$}
& \small{\emph{(e)} zoom $sg^{mrk}(\varrho^{\chi^{2}}(\mathbf{f}_{\lambda}),\widehat{\kappa})$}\\
\includegraphics[width=0.25\columnwidth]{roujan_N50_M100_S10_Rmax30_mpdf_marqueurs_boules_conditioned_classif_clara_3classes_axesAFC_1_3_bis.png} &
\includegraphics[width=0.25\columnwidth]{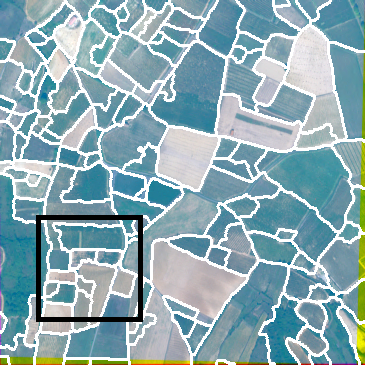} &
\includegraphics[width=0.25\columnwidth]{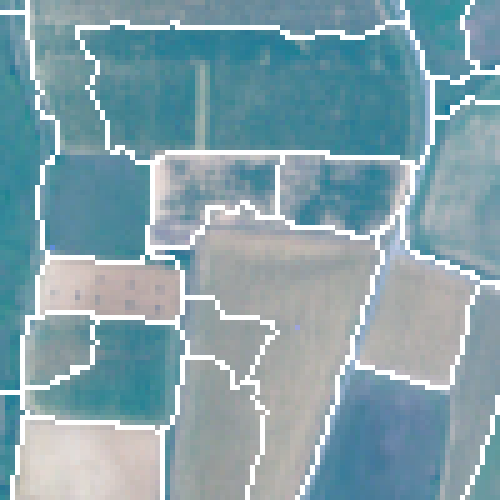}\\
\small{\emph{(f)} $mpdf^{\kappa-b}(\mathbf{f}_{\lambda})$} &
\small{\emph{(g)} $sg^{mrk}(mpdf^{\kappa-b}(\mathbf{f}_{\lambda}) ,
\widehat{\kappa})$} & \small{\emph{(h)} zoom $sg^{mrk}(mpdf^{\kappa
-b}(\mathbf{f}_{\lambda}) , \widehat{\kappa})$}\\
&  &
\end{tabular}
\end{tabular}
\caption{(First line) \emph{(a)} Classification in 3 classes
$\kappa$ and \emph{(b)} markers coming from the classification
$\widehat{\kappa}$ on image ``Roujan". (Second line) \emph{(d, e)}
deterministic approach
$sg^{mrk}(\varrho^{\chi^{2}}(\mathbf{f}_{\lambda
}),\widehat{\kappa})$ on \emph{(c)} the chi-squared metric based gradient $\varrho^{\chi^{2}%
}(\mathbf{f}_{\lambda})$. (Third line) \emph{(g, h)} Stochastic approach $sg^{mrk}%
(mpdf^{\kappa-b}(\mathbf{f}_{\lambda}) , \widehat{\kappa})$ on
\emph{(f)} the marginal pdf $mpdf^{\kappa-b}(\mathbf{f}_{\lambda})$.
The parameters are $Q=3$ classes, $N=50$ germs $M=100$ realisations,
minimal area $S=10$ pixels, maximum radius $Rmax=30$ pixels). For
presentation of images without zoom, the contours are
dilated by a SE $3 \times3$ pixels.}%
\label{Fig_segmentation_roujan}%
\end{figure}

\begin{figure}
\centering
\begin{tabular}
[c]{ccc}%
\includegraphics[width=0.25\columnwidth]{portdebouc_classif_clara_7classes_axesAFC_1_2.png} &
\includegraphics[width=0.25\columnwidth]{portdebouc_marqueurs_classif_fond_noir.png} &
\\
\small{\emph{(a)} $\kappa$} &
\small{\emph{(b)} $\widehat{\kappa}$} & \\
&  &
\end{tabular}
\newline%
\begin{tabular}
[c]{@{}c@{ }c@{ }c@{}}%
\includegraphics[width=0.25\columnwidth]{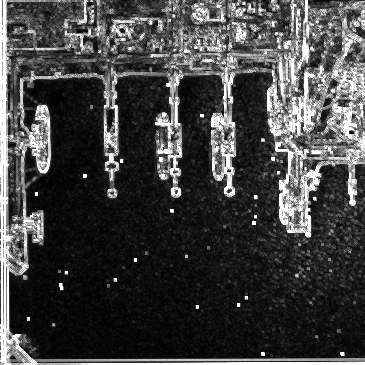} &
\includegraphics[width=0.25\columnwidth]{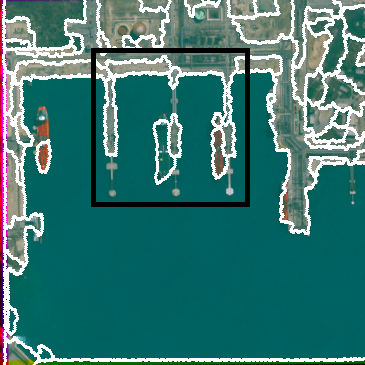} &
\includegraphics[width=0.25\columnwidth]{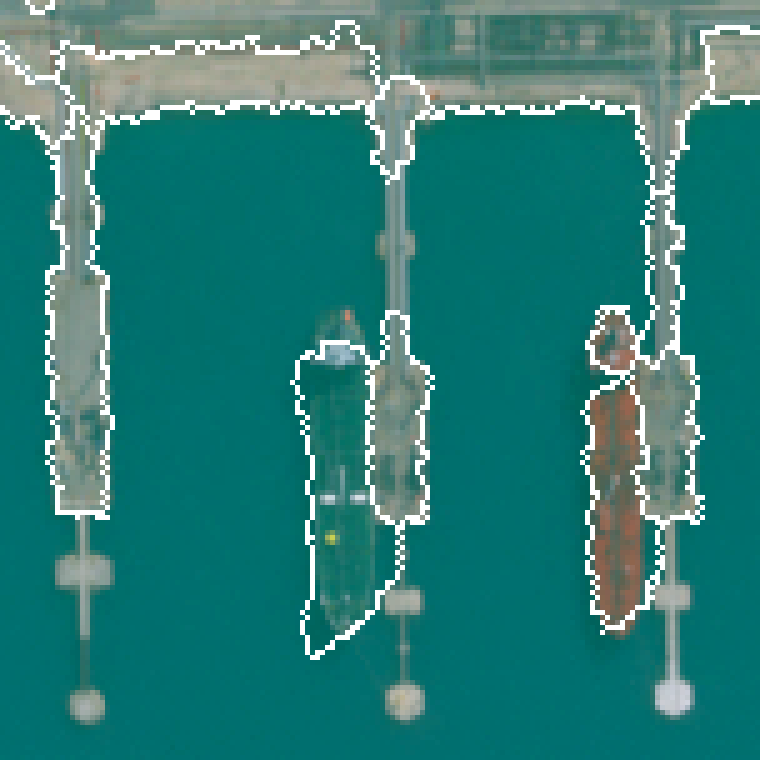}\\
\small{\emph{(c)} $\varrho^{\chi^{2}}(\mathbf{f}_{\lambda})$} &
\small{\emph{(d)}
$sg^{mrk}(\varrho^{\chi^{2}}(\mathbf{f}_{\lambda}),\widehat{\kappa})$}
& \small{\emph{(e)} zoom $sg^{mrk}(\varrho^{\chi^{2}}(\mathbf{f}_{\lambda}),\widehat{\kappa})$}\\
\includegraphics[width=0.25\columnwidth]{portdebouc_N100_M100_S10_Rmax30_mpdf_marqueurs_boules_conditioned_classif_clara_7classes_axesAFC_1_2_bis.png} &
\includegraphics[width=0.25\columnwidth]{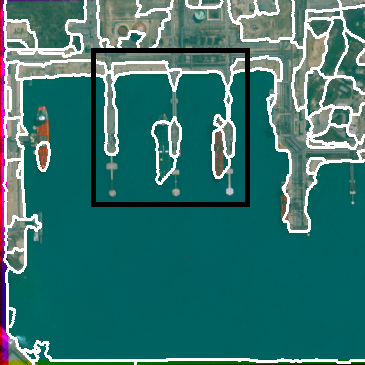} &
\includegraphics[width=0.25\columnwidth]{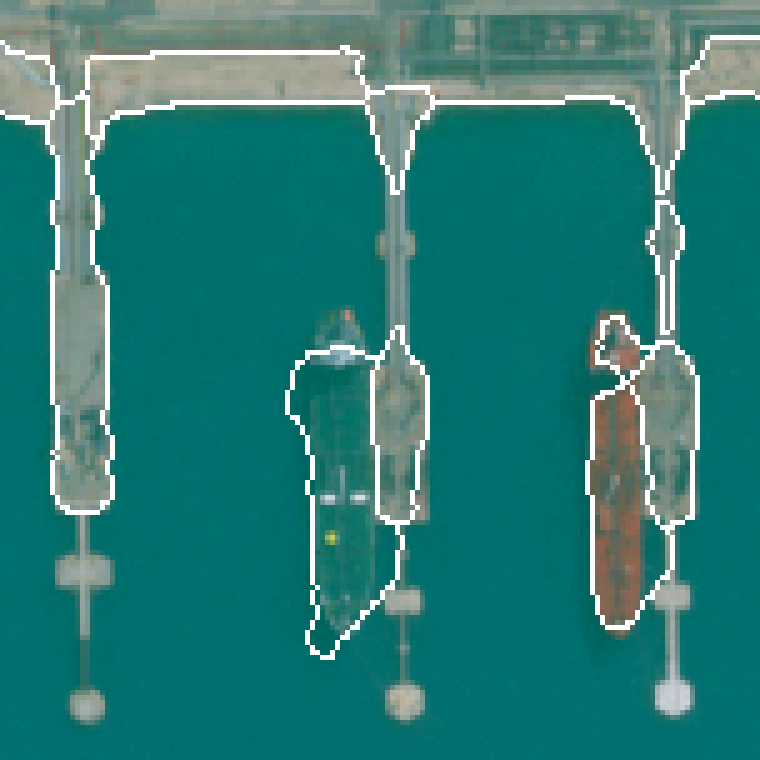}\\
\small{\emph{(f)} $mpdf^{\kappa-b}(\mathbf{f}_{\lambda})$} &
\small{\emph{(g)} $sg^{mrk}(mpdf^{\kappa-b}(\mathbf{f}_{\lambda}) ,
\widehat{\kappa})$} & \small{\emph{(h)} zoom $sg^{mrk}(mpdf^{\kappa
-b}(\mathbf{f}_{\lambda}) , \widehat{\kappa})$}\\
&  &
\end{tabular}
\caption{First line) \emph{(a)} Classification in 3 classes $\kappa$
and \emph{(b)}  markers coming from the classification
$\widehat{\kappa}$ on image ``Port de Bouc". (Second line) \emph{(d,
e)} deterministic approach
$sg^{mrk}(\varrho^{\chi^{2}}(\mathbf{f}_{\lambda
}),\widehat{\kappa})$ on \emph{(c)} the chi-squared metric based gradient $\varrho^{\chi^{2}%
}(\mathbf{f}_{\lambda})$. (Third line) \emph{(g, h)} Stochastic approach $sg^{mrk}%
(mpdf^{\kappa-b}(\mathbf{f}_{\lambda}) , \widehat{\kappa})$ on
\emph{(f)} the marginal pdf $mpdf^{\kappa-b}(\mathbf{f}_{\lambda})$.
The parameters are $Q=7$ classes, $N=50$ germs $M=100$ realisations,
minimal area $S=10$ pixels, maximum radius $Rmax=30$ pixels). For
presentation of images without zoom, the contours are
dilated by a SE $3 \times3$ pixels.}%
\label{Fig_segmentation_port_de_bouc}%
\end{figure}

The parameters for the stochastic WS controlled by markers are the
same as in part \ref{part_seg_pdf}. In figure
\ref{Fig_segmentation_roujan}, are given the results of segmentation
by deterministic and stochastic WS for image ``Roujan" with markers
coming from the transformed classification $\widehat{\kappa}$. We
observe that the contours are smoother and follow more precisely the
main limits of the regions for the stochastic approach than for the
deterministic one. The same observation is made on the image ``Port
de Bouc" classified in $Q=7$ classes. (fig.
\ref{Fig_segmentation_port_de_bouc}).

\FloatBarrier

\subsection{General method of multispectral segmentation by stochastic WS}

\begin{figure}
\centering
\includegraphics[width=0.6\columnwidth]{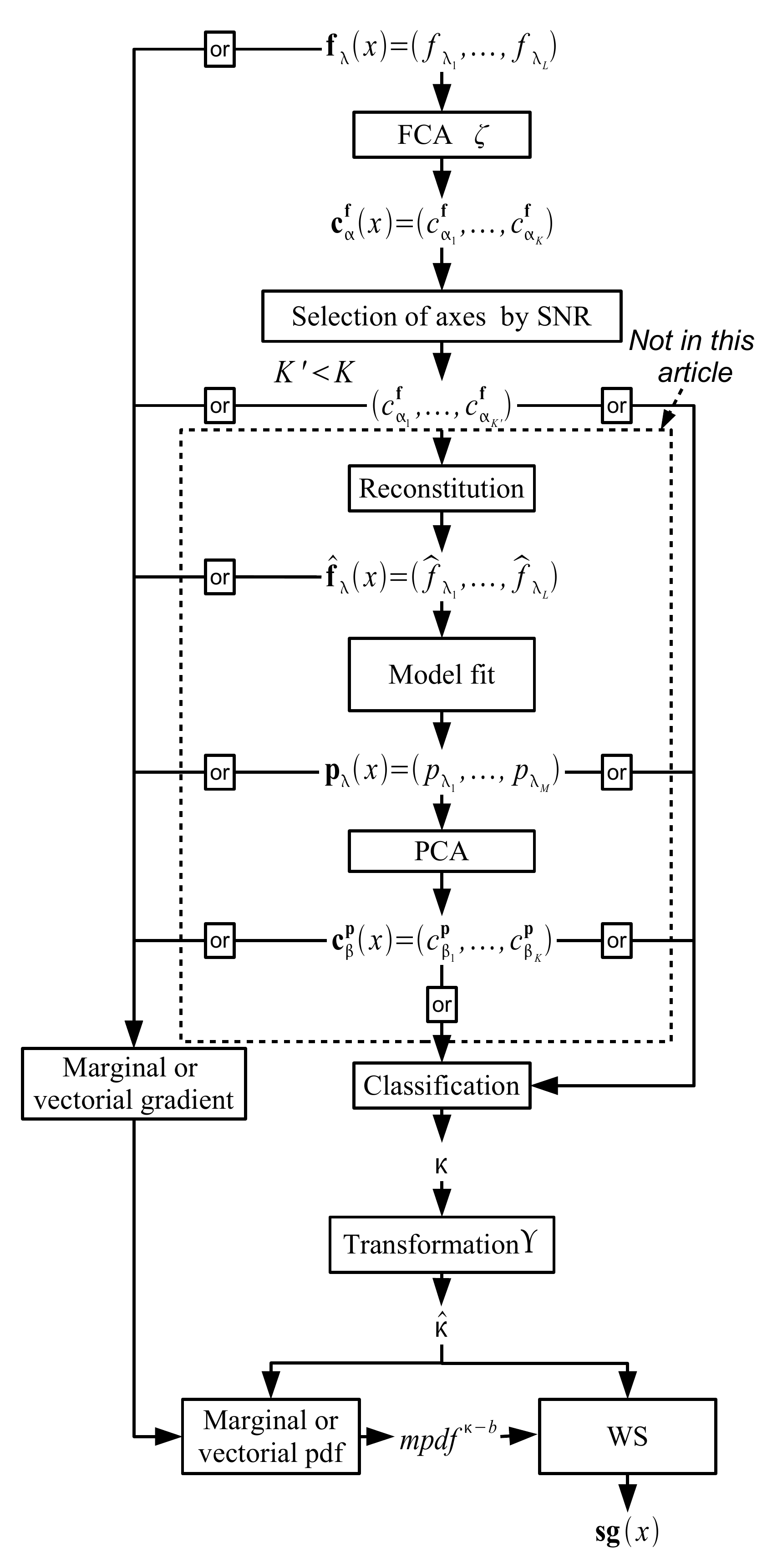}
\caption{General framework of WS controlled by a classification,
with regionalized random-balls germs. The part outlined with a
dotted line is not
presented in this paper.}%
\label{sch_gen_seg}%
\end{figure}

After all these observations and analysis, we can summarize now the
algorithmic pipeline of the general method of segmentation of
multispectral image by stochastic WS. The framework is presented on
figure \ref{sch_gen_seg}. The different steps are as follows:

\begin{itemize}
\item a FCA to reduce the spectral dimension of the image $\mathbf{f}_{\lambda}$
and eventually to  filter the noise. This transformation creates a
factor image  $\mathbf{c}^{\mathbf{f}}_{\alpha}$

\item the previous step requires the selection of factorial axes by the signal
to noise ratio (SNR) of the factor pixels

\item a spectral classification to group pixels into homogeneous classes
(global point-wise comparison)

\item a morphological transform $\Upsilon$ of each classes. The image
transformed is noted $\widehat{\kappa}=\Upsilon(\kappa)$

\item the building of a pdf of contours that can be marginal $mpdf$ or
vectorial $vpdf$ starting from a gradient on the image space
$\mathbf{f}_{\lambda}$ (MIS) or factor space
$\mathbf{c}_{\alpha}^{\mathbf{f}}$ (FIS) with random-balls germs
regionalized by the spectral classification

\item a segmentation of the pdf by a WS controlled by markers coming from the
classes of the transformed classification $\widehat{\kappa}$
\end{itemize}

Some of the steps presented in the flowchart of figure
\ref{sch_gen_seg} have not been discussed in this paper. These steps
are very useful for other applications of multivariate image
segmentation; the interested reader can find more details in
\citet{Noyel_ISBI_2008}.

\FloatBarrier

\section{Conclusion}
\label{sec_conclusion}

A novel method to segment multispectral images is presented in this
paper. First, a dimensionality reduction by Factor Correspondence
Analysis is performed. It reduces the spectral noise and preserves
the spatial information. A new way to select factor axes is
presented. It uses the spatial information carried by the factor
pixels (i.e. signal to noise ratio) and not on only the statistical
information (i.e. inertia). Then, a classification stage ensures a
global pixel-wise comparison and groups pixels into classes with
similar spectra. A morphological transformation is subsequently
applied to obtain markers for the final segmentation. Then, a
probability density function of contours conditioned by the previous
classification is built with random ball-germs. This pdf contains
spatial and spectral information and is estimated by Monte-Carlo
simulations. During the pdf estimation process regional information
between pixels is compared. Finally, the pdf is segmented by a WS
controlled by markers coming from the transformed classification.

Moreover, this spatio-spectral method of segmentation needs few and well
controlled parameters:
\begin{itemize}
\item the threshold of the signal to noise ratio for the Factor Correspondence
Analysis (dimensionality reduction stage); this can be selected
according to the knowledge of the amount of noise in the images;

\item the number of classes of the classification which corresponds to the
number of different kinds of spectra  in the image. This parameter
is more robust than the number of regions with the largest volume in
the standard hierarchical watershed segmentation;

\item eventually, a size criterion to remove small regions which
must be excluded of the segmentation;

\item the number of random germs (linked to the number of  regions)
$N$;

\item the number of realisations (fixed) $M = 100$;

\item the maximum radius of the ball (related to the surface area of interest zones) $Rmax$.
\end{itemize}

It is shown, that in the case of segmentation of similar images
these parameters can be easily  fixed.

The approach is perfectly valid for VHR images: the only requirement
is to start from an initial classification representing all the most
significant regions, and of course, the regions must have a minimal
thickness to define the notion of inner maker for the watershed
segmentation.

This general method of segmentation is not limited to the field of
remote sensing but can also be applied on medical images
\citep{Noyel_ISBI_2008}, microscopy images, thermal images
\citep{Noyel_IAS_2007}, time series, multivariate series, etc.

We are now thinking about two interesting perspectives: to introduce
more prior information on the construction of the pdf of contours
and to use a more advanced multivariate kernel during the estimation
process of the pdf.

\section{Acknowledgments}

The authors are grateful to Prof. Guy Flouzat (Laboratoire de
T\'{e}l\'{e}d\'{e}tection \`{a} Haute R\'{e}solution, LTHR/ ERT 43 /
UPS, Universit\'{e} Paul Sabatier, Toulouse 3, France) for providing
the PLEIADES satellite simulated images, obtained in the framework
of ORFEO program (Centre National d'Etudes Spatiales, the French
space agency).

The authors would like to dedicate this paper to the memory of Prof.
Guy Flouzat.

%pour que WinEdt trouve le fichier de biblio
%GATHER{tRESguide.bib}
\bibliographystyle{tRES}
\bibliography{tRESguide}

\end{document}